\documentclass[runningheads]{llncs}

\PassOptionsToPackage{table,dvipsnames}{xcolor}

\usepackage{eccv}

\usepackage{eccvabbrv}

\usepackage{graphicx}
\usepackage{booktabs}
\usepackage{enumitem}
\usepackage{caption}

\usepackage{colortbl}

\usepackage[accsupp]{axessibility}  

\usepackage{booktabs}
\usepackage{siunitx}         

\usepackage{array}
\usepackage{adjustbox}

\usepackage{wrapfig}

\usepackage{algorithm}
\usepackage{algpseudocode}

\definecolor{tblborder}{RGB}{210,214,220}   
\definecolor{tblhead}{RGB}{244,246,249}     
\definecolor{tblstripe}{RGB}{250,251,253}   
\definecolor{tblsection}{RGB}{236,242,252}  
\definecolor{tblhighlight}{RGB}{225,240,255}

\usepackage{hyperref}

\usepackage{orcidlink}


\usepackage[most]{tcolorbox}
\tcbuselibrary{listings,breakable}

\newtcblisting{promptbox}[1][]{
  breakable, enhanced, colback=gray!5, colframe=black!70,
  boxrule=0.5pt, arc=2pt, left=6pt, right=6pt, top=4pt, bottom=4pt,
  fonttitle=\bfseries, listing only,
  listing options={
    basicstyle=\ttfamily\footnotesize,
    breaklines=true, breakatwhitespace=true,
    columns=fullflexible, keepspaces=true
  }, #1
}

\begin{document}

\title{Reasoning-Guided Part-Level Visual Grounding via Reinforcement Learning} 

\titlerunning{Reasoning-Guided Part-Level Visual Grounding}

\author{Kazi Sajeed Mehrab \and
Hani Alomari \and
Najibul Haque Sarker \and
Chia-Wei Tang \and
Zaber Ibn Abdul Hakim \and
Anuj Karpatne \and
Chris Thomas}

\authorrunning{K. Mehrab et al.}

\institute{Department of Computer Science\\Virginia Tech\\
\email{\{ksmehrab, hani, najibulhaque, cwtang, zaberhakim, karpatne, christhomas\}@vt.edu}
}

\maketitle

\begin{center}
\captionsetup{type=figure}
\includegraphics[width=\linewidth]{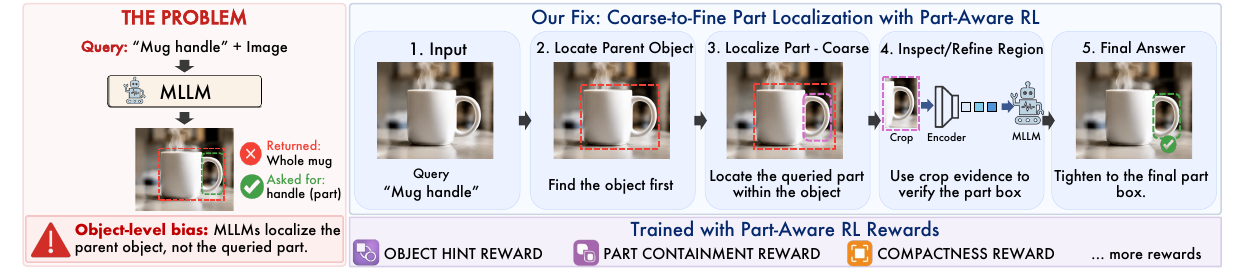}
\caption{Our method grounds parts coarse-to-fine: locate the object, then the part within it, then self-checks, re-encoding the predicted crop to refine.}
\label{fig:overview}
\end{center}

\begin{abstract}
Multimodal large language models (MLLMs) ground whole objects well from free-form language queries, but they struggle when the query names a part rather than the object. We trace this to a missing object-part hierarchy, since parts are localized in the same single step used for objects. We propose Object-Part Hierarchical Reflective Grounding (OP-HRG), a coarse-to-fine reasoning-guided grounding strategy that first localizes the parent object and then the part within it. A self-check then reflects on the result, with an extension to re-encode the predicted crop to inspect the region it is correcting. We introduce a part-aware GRPO framework to train our pipeline with stage-wise rewards. 
A 4B model trained this way outperforms 7B grounding LLMs and SAM3 across PascalPart, PartImageNet, and InstructPart, and transfers to reasoning segmentation. \footnote{Code and models will be available at \url{https://github.com/sajeedmehrab/op-hrg}.}
\end{abstract}
\section{Introduction}
\label{sec:intro}

Localizing the parts of an object is a core requirement in many real-world settings. A robot told to ``grasp the mug (object) by its handle (part)'' must distinguish the handle from the body, and a clinician reading a scan must isolate specific anatomical structures (parts) rather than whole organs (objects). We use \emph{part} in the sense established by part-segmentation benchmarks \cite{chen2014detect, wei2023ov, he2022partimagenet, instructpart}: a constituent sub-region of a parent object -- such as a car's wheel or a mug's handle -- that is defined relative to the whole object and is often tied to a specific function or affordance. Localizing a part is harder than localizing the object that contains it: rather than finding a self-contained entity, a model must reason about spatial layout and containment relative to the parent object, much as a person first locates the mug and only then its handle.

Multimodal large language models (MLLMs) such as Qwen-VL \cite{wang2024qwen2, bai2025qwen3} perform this localization through \emph{visual grounding} -- producing a box or mask for the image region named by a free-form query. The strongest publicly available, widely used MLLMs are now strong zero-shot object grounders, yet the same models remain weak when the query names a part rather than a whole object. When asked to ground a mug's handle, an MLLM is likely to return the entire mug instead (Figure~\ref{fig:overview}). Two factors underlie this gap. First, vision-language pretraining is dominated by object-level descriptions, with part annotations comparatively rare and costly to obtain \cite{he2022partimagenet, instructpart, wei2023ov}, biasing models toward coarse entities. Second, grounding MLLMs typically handle every query the same way. A part query goes through the same single-step localization as an object, with no mechanism to exploit the natural hierarchy between an object and its parts. 

Prior part-grounding approaches span supervised segmenters, token-based MLLMs, and decoupled MLLM-to-SAM pipelines (Sec. \ref{sec:related_work}). The closest to ours, Seg-Zero \cite{segzero} and VisionReasoner \cite{visionreasoner}, ground whole objects well, but like other grounding MLLMs they localize every query in a single step with no signal for object-part reasoning. We posit that MLLMs already have the capacity for hierarchical reasoning, but it stays dormant without a structured way to activate it, and reinforcement learning offers a direct way to reward and strengthen object-part reasoning. 

We therefore propose Object-Part Hierarchical Reflective Grounding (OP-HRG), a prompting paradigm that guides MLLMs to \emph{observe, reason, and localize in a structured coarse-to-fine manner}. Under OP-HRG, the model first decides whether the query refers to an object or a part. If it is a part, the model localizes the parent object as an anchor, then generates an initial part localization within that anchor. As a complementary step, the model reflects on its own answer, checking whether an adjustment is needed before producing the final result. This step-by-step structure mirrors how careful human observation works: locate the object, focus on the part, then verify. We pair this prompting paradigm with a dedicated part-aware Group Relative Policy Optimization (GRPO) framework. Unlike prior RL-based grounding methods that optimize a single localization reward, our reward provides separate, verifiable signals for each stage of the OP-HRG output: parent-object and part localization accuracy, part-in-object containment, self-reflection consistency, and improvement of the final answer over the initial one.

We evaluate our approach in the cross-dataset zero-shot setting on PascalPart \cite{chen2014detect, wei2023ov} and PartImageNet \cite{he2022partimagenet} -- benchmarks that part-specific segmenters typically train on, so we compare against MLLM-grounding methods that typically share our zero-shot setting. Using the Qwen3-VL-Instruct-4B model as our backbone, smaller than the 7B models used by competing methods, our method outperforms these baselines on both object and part categories. Ablation studies confirm that both the OP-HRG prompting and the part-aware rewards are necessary. Our contributions are:

\begin{itemize}[leftmargin=*,nosep]
    \item We analyze the part-grounding gap in MLLMs and show that current pipelines, including RL-optimized ones, underperform on part queries.
    \item We propose OP-HRG, a structured prompting paradigm that guides MLLMs through object-first hierarchical localization, complemented by a self-reflective step and an active visual perception extension that re-encodes predicted regions as visual evidence for the critique. Our analysis shows the self-reflective step acts mainly as a train-time regularizer.
    \item We introduce a part-aware GRPO reward framework with stage-wise, verifiable rewards covering parent-object accuracy, part containment, critique consistency, and answer improvement.
    \item Across three benchmarks, our 4B-parameter model improves part grounding over strong MLLM-grounding baselines -- cross-dataset zero-shot on PascalPart and PartImageNet, and in-domain on InstructPart, scales to an 8B backbone, shows only a modest trade-off in general object referring (RefCOCO/+/g), and transfers to reasoning segmentation (ReasonSeg).
\end{itemize}
\section{Related Work}
\label{sec:related_work}

\noindent \textbf{Part-Level Segmentation.}
Part segmentation has traditionally been approached through supervised methods trained on fixed label sets, evaluated on benchmarks such as PartImageNet \cite{he2022partimagenet}, PascalPart \cite{chen2014detect}, and InstructPart~\cite{instructpart}. Open-vocabulary extensions \cite{sun2023going, li2024partglee, choi2025fine} generalize to unseen parts through cross-modal correspondence learning or cost aggregation, while unified formulations such as Semantic-SAM \cite{li2024segment} unify object-level and part-level predictions within a single framework. However, these approaches rely on part-specific segmentation supervision or learned visual–semantic correspondences, limiting their flexibility. Our work is different from this paradigm: rather than training a pixel-level part-segmentation model, we elicit part localization through structured MLLM reasoning and reinforcement learning with dedicated part-aware rewards. We target part grounding within MLLMs rather than part segmentation in general.

\noindent \textbf{Promptable Segmentation Models.} The Segment Anything Model (SAM) \cite{sam} established promptable segmentation at scale, producing high-quality masks from spatial prompts such as points and boxes. SAM2 \cite{sam2} improves upon SAM-v1 for image segmentation from spatial prompts, and SAM3 \cite{sam3} adds text-conditioned segmentation, predicting masks directly from short phrases. Because these models decode accurate masks from lightweight prompts, they serve as a common mask decoder in the MLLM-based grounding methods discussed below.

\noindent \textbf{Visual Grounding with MLLMs.}
Recent MLLMs have enabled zero-shot visual grounding by localizing image regions from natural-language queries. Special-token methods such as LISA \cite{lisa}, GLaMM \cite{rasheed2024glamm}, Sa2VA \cite{yuan2025sa2va}, PixelLM \cite{ren2024pixellm}, and UniPixel \cite{liu2026unipixel} embed segmentation tokens into the LLM output to condition a jointly trained mask decoder, the dominant MLLM-grounding paradigm. Referring frameworks \cite{yuan2024osprey, you2023ferret} and open-set detectors such as Grounding DINO \cite{liu2024grounding} and Grounded SAM \cite{ren2024grounded} further advance visual grounding. More recently, reinforcement learning has emerged as an effective alignment strategy for grounding MLLMs. GRPO \cite{shao2024deepseekmath} removes the critic network to improve training efficiency and has been widely adopted in visual reasoning \cite{liu2025visual, shen2025vlm, yang2025r1, chen2025suitability, meng2025mm} and visual grounding \cite{you2025seg, he2026dr, zhu2025lens, bai2025univg, cao2025ground, zhou2026affordancegrasp, shen2025satori}, typically with IoU-oriented rewards. Most closely related to us are the decoupled MLLM-to-SAM pipelines Seg-Zero \cite{segzero} and VisionReasoner \cite{visionreasoner}, which pair GRPO-based optimization with a frozen mask decoder for reasoning-aware segmentation -- the same paradigm we adopt. However, both treat all queries uniformly and optimize a single-stage localization reward, without modeling the object-part hierarchy or providing part-specific reward signals. Our framework extends this line of work by introducing object-part hierarchical reasoning and dedicated part-aware rewards into the GRPO alignment process.

\noindent \textbf{Chain-of-Thought and Structured Reasoning for Visual Tasks.}
Multimodal chain-of-thought methods \cite{zhang2023multimodal, Mitra_2024_CVPR, zhang-etal-2025-improve, fu2025refocus} improve reasoning through intermediate rationales, and spatially grounded approaches \cite{Chen_2024_CVPR, man2025argus, li-etal-2025-vocot, sharma2025think} align reasoning steps with image regions. Reflection-oriented paradigms \cite{jian2025lookagainthinkslowly, ma2025deepperception} add self-critique to improve model's behavior and final answer quality. Our OP-HRG strategy builds on these directions: it enforces spatially grounded hierarchical reasoning through object-first localization and combines it with a self-reflective check for part-level grounding refinement and verification.
\section{Method}
\label{sec:method}

\begin{figure*}[t]
  \centering
  \includegraphics[width=1\linewidth]{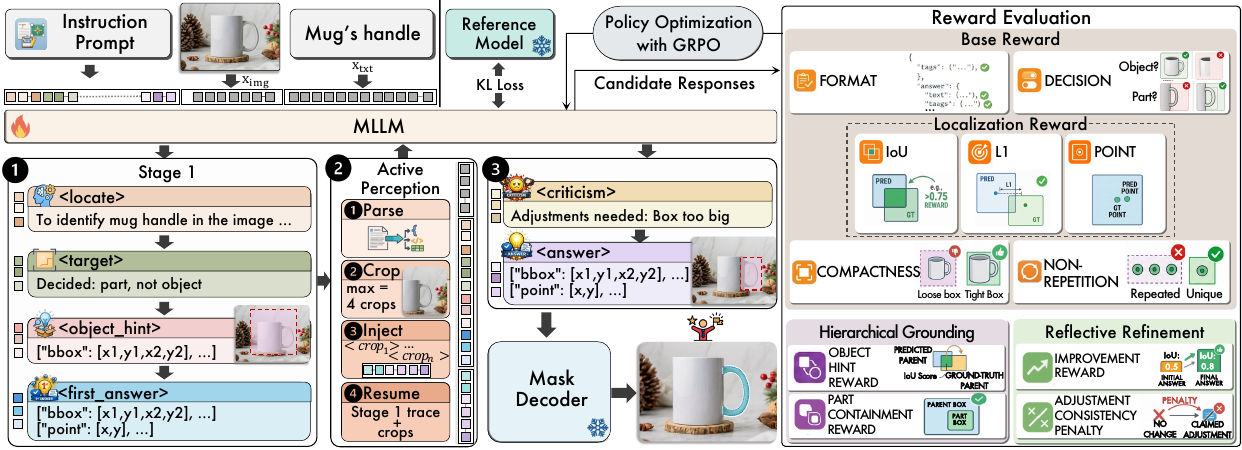}
   \caption{Overview of the action steps and reward evaluation for OP-HRG.}   
   \label{fig:method_figure}
   \vspace{-1.5em}
\end{figure*}

In this section, we describe our model-agnostic reinforcement learning framework for grounding objects and object parts using the reasoning capabilities of multimodal LLMs (MLLM). Our approach combines a structured object-part prompting paradigm, which we term Object-Part Hierarchical Reflective Grounding (OP-HRG), with a part-aware reinforcement learning objective based on Group Relative Policy Optimization (GRPO). Figure \ref{fig:method_figure} gives an overview of the action steps and reward evaluation.

\subsection{Problem Formulation}

We consider the task of \textbf{visual grounding of objects and object parts}. 
Our system receives an RGB image $I \in \mathbb{R}^{H \times W \times 3}$ and a natural language query 
$q$. The query may denote a whole object category (e.g., ``cat'') or a semantic part of an object 
(e.g., ``cat's tail''). 
During inference, the model must interpret the image and the query and return spatial localizations for the 
visual entity described by $q$, even when the object or part name has not been observed during training. Specifically, our end goal is to predict a binary segmentation mask that delineates $q$.

\subsection{Decoupled Reasoning and Segmentation Architecture}

Recent multimodal LLMs, like Qwen-VL \cite{wang2024qwen2, bai2025qwen3}, demonstrate strong reasoning and coordinate generation abilities, while large vision models such as the Segment Anything Model (SAM) \cite{sam, sam2, sam3} predict accurate, dense segmentation masks from geometric prompts. This complementarity motivates separating semantic reasoning from pixel segmentation, a decoupled design also adopted by recent grounding MLLMs \cite{lisa, yuan2025sa2va, ren2024pixellm, liu2026unipixel, rasheed2024glamm, segzero, visionreasoner}.
We adopt this design as well.

In our pipeline, the MLLM reads $(I,q)$ and generates bounding boxes and representative points 
$\mathcal{O}$. These localization primitives are then passed, together with the image, to a mask decoder, which outputs a segmentation mask $S \in \{0,1\}^{H \times W}$. Concretely, for each input $(I,q)$, the MLLM produces a \textbf{set of localization primitives}
\begin{equation}
\mathcal{O} = \{(b_i,p_i)\}_{i=1}^{K},
\end{equation}
where $K$ is the number of predicted regions matching the query. 
Each $b_i = (x_{1}^{i},y_{1}^{i},x_{2}^{i},y_{2}^{i})$ represents a tight 2D bounding box in image coordinates 
with $0 \leq x_{1}^{i} < x_{2}^{i} \leq W$ and $0 \leq y_{1}^{i} < y_{2}^{i} \leq H$. 
The primitive $p_i=(x^{i},y^{i})$ is a representative point required to lie inside the queried entity. 

\begin{figure}[t]
  \centering
  \includegraphics[width=1\linewidth]{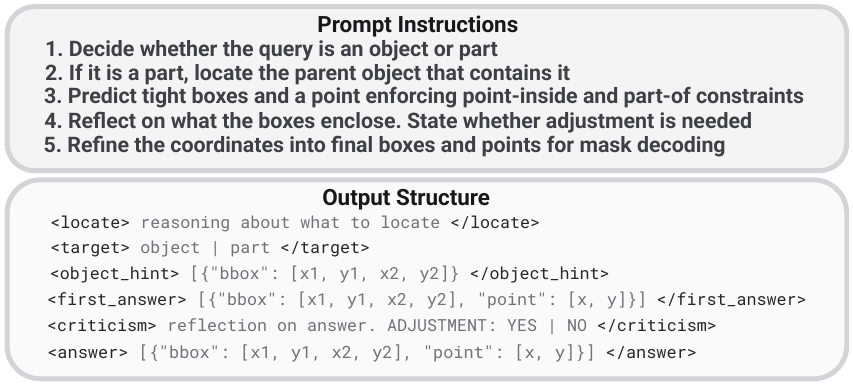}
   \caption{High-level structure of OP-HRG prompt and corresponding output format.}
   \vspace{-1.5em}
   \label{fig:prompt_figure}
\end{figure}

\subsection{Object-Part Hierarchical Reflective Grounding}

To address the challenges of grounding object parts, we introduce Object-Part Hierarchical Reflective Grounding (OP-HRG), a structured prompting paradigm that explicitly encodes the object-part hierarchy and incorporates self-reflective refinement. Figure \ref{fig:prompt_figure} illustrates the high-level structure of the OP-HRG prompt and the corresponding expected output format. The full prompt and example outputs are in the Appendix \ref{app:prompt}.  

\textit{First}, the model reasons about the query and determines whether it refers to a whole object or a part. This early decision governs the subsequent steps and enforces an explicit distinction between object-level and part-level grounding. \textit{Second}, when the query refers to a part, the model localizes the parent object before the part itself, enforcing a coarse-to-fine process in which part localization is conditioned on object-level context, mirroring how humans search for object parts. \textit{Third}, the model produces an initial localization as bounding boxes and representative points; for parts, these must respect the structural constraint that part boxes lie within the predicted object boxes. \textit{Finally}, OP-HRG incorporates a self-reflective verification step in which the model critiques its own initial prediction, assesses whether the localization is tight and accurate, and decides whether adjustment is necessary, then either retains the initial prediction or produces a refined one. During training, this reflect-and-revise step provides a complementary corrective check on the initial localization.

As shown in Figure \ref{fig:prompt_figure}, the prompt enforces this reasoning as a fixed sequence of tagged outputs that expose intermediate reasoning states and localization decisions (reasoning trace, target decision, object hint, initial answer, critique, and refined answer). This structured format serves two purposes. First, it induces the desired hierarchical reasoning behavior during inference. Second, it enables fine-grained, verifiable reward signals for our reinforcement learning framework.


\subsection{Active Visual Perception}
\label{sec:active_perception}
In the OP-HRG formulation above, the self-reflective step reasons only over the textual bounding-box and point coordinates of the initial prediction. The reflective step therefore needs to resolve these coordinates against the whole image, rather than being able to inspect the enclosed image region of the model's first answer directly. We therefore extend OP-HRG with an \emph{active visual perception} (AP) step that grounds the reflective step in fresh visual evidence. After the initial localization (\texttt{<first\_answer>}), generation pauses and the predicted boxes are used to crop the corresponding regions from the input image. Each crop is re-encoded by the model's pretrained vision encoder and injected back as interleaved visual tokens, so the model performs its critique and reaches the final answer (\texttt{<criticism>} and \texttt{<answer>}) on this fresh visual evidence. This loop is applied both during training and at inference, so the policy learns to exploit the re-encoded crops when deciding if it needs to adjust its prediction.

\subsection{Reinforcement Learning with Part-Aware Rewards}
\label{sec:reward_design} 
Although pretrained MLLMs possess substantial visual and semantic knowledge, they exhibit a bias toward coarse object-level grounding, in part due to the scarcity of part-centric supervision in large-scale MLLM training \cite{instructpart, he2022partimagenet, liu2023partslip}. Reinforcement learning offers a potential mechanism for correcting this bias -- it lets us reward tight, compact localizations and enforce structural constraints such as part–object containment and consistency between intermediate reasoning and final outputs. While prior work shows GRPO \cite{shao2024deepseekmath} is well-suited for aligning instruction-tuned MLLMs for visual grounding \cite{segzero, visionreasoner}, these models remain weak at grounding object parts, motivating a dedicated framework for part-centric localization and object-part reasoning. We therefore train our multimodal LLM with GRPO under a composite, part-aware reward, using its reasoning to produce more accurate and compact object- and part-level bounding boxes and representative points. The reward is built from modular, verifiable components in three groups: \textbf{base rewards}, \textbf{hierarchical grounding rewards}, and \textbf{reflective refinement rewards}, described below. All components are normalized to $[0,1]$ (or $[-1,0]$ for penalties) and combined into a total reward normalized by the maximum achievable score (full normalization in Appendix~\ref{app:reward}). We detail the GRPO objective in Section~\ref{subsec:training_objective} and the detailed reward implementation in Appendix~\ref{app:reward}.

\paragraph{\textbf{Base Rewards.}}
These rewards apply to every query regardless of whether it targets an object or a part.

The \textit{format compliance reward} scores adherence to the OP-HRG tag sequence and the validity of JSON content within each tag, assigning individual credit for well-formed \texttt{<object\_hint>}, \texttt{<first\_answer>}, \texttt{<answer>}, and \texttt{<criticism>} blocks, normalized by the maximum achievable format score. The \textit{target decision reward} is a binary signal that rewards correct classification of the query as \textit{object} or \textit{part}, which determines whether the model activates the object-part hierarchical reasoning path.

The \textit{localization rewards} assess the geometric quality of predicted bounding boxes and representative points, computed identically for the initial and final predictions (\texttt{<first\_answer>} and \texttt{<answer>} respectively) to provide signal at both stages. When multiple instances are present, we construct a cost matrix $C = \mathbf{1} - \mathrm{IoU}(\hat{\mathcal{B}}, \mathcal{B}^{*})$ between the $M$ predicted and $N$ ground-truth boxes and solve the assignment via the Hungarian algorithm~\cite{kuhn1955hungarian}, scoring matched pairs and averaging over $\max(M,N)$ to penalize both missing and spurious detections. We reward three localization components:
\begin{itemize}[leftmargin=*,nosep]
    \item \textit{IoU Reward}: the mean IoU over matched pairs.
    \item \textit{L1 Distance Reward}: for each matched pair $(i,j)$, the mean absolute coordinate difference $\ell_{1,ij}$ under an adaptive threshold $\tau^{(\ell_1)}_{j} = \alpha_{\ell_1} \cdot d_j$, where $d_j$ is the diagonal of the $j$-th ground-truth box, $\alpha_{\ell_1}$ a scaling factor, and the threshold is clamped to $\tau^{(\ell_1)}_{\max}$, giving the per-pair reward $\max(0,\, 1 - \ell_{1,ij}/\tau^{(\ell_1)}_{j})$.
    \item \textit{Point Accuracy Reward}: a predicted point receives credit only if it lies inside its own predicted box \emph{and} its distance to the matched ground-truth point falls within $\tau^{(p)}_{j} = \alpha_{p} \cdot d_j$, similarly clamped to $\tau^{(p)}_{\max}$.
\end{itemize}

Scaling the tolerance by the ground-truth box diagonal judges accuracy relative to region size, which matters for small part boxes that a fixed threshold would treat too permissively. We use a tighter scaling factor for box coordinates ($\alpha_{\ell_1}$) than for representative points ($\alpha_{p}$), since a point may lie anywhere within the target region whereas box boundaries must align tightly with ground truth.

The \textit{compactness reward} encourages tight spatial coverage by scoring each Hungarian-matched pair $(i,j)$ with two complementary terms over the predicted box $\hat{b}_i$ and its matched ground-truth box $b^*_j$: a \textit{precision} term $\rho_{ij} = |\hat{b}_i \cap b^*_j| \,/\, |\hat{b}_i|$, the fraction of the predicted box overlapping ground truth, and an \textit{over-prediction penalty} $\omega_{ij} = -\min\!\bigl(1,\, \max(0,\, |\hat{b}_i|/|b^*_j| - 1)\bigr)$, saturating at $-1$ once the predicted box reaches twice the ground-truth area. The reward is the mean of $\rho_{ij} + \omega_{ij}$ across matched pairs, favoring tight boxes; this benefits object parts in particular, where even modest over-prediction can encompass neighboring parts or the parent object.

The \textit{non-repetition reward} penalizes degenerate chain-of-thought outputs by checking for repeated sentences in the reasoning trace. The reward drops to zero if two or more exact sentence duplicates are detected.

\paragraph{\textbf{Hierarchical Grounding Rewards.}}
These components are active only for part queries and directly correspond to the object-part hierarchy introduced by OP-HRG.

The \textit{object hint reward} evaluates the predicted parent-object boxes, which are matched to ground-truth object boxes via Hungarian matching on IoU, and the reward is the normalized average IoU over matched pairs.

The \textit{part containment reward} enforces structural constraints on the final part predictions. For each predicted part box $\hat{b}^{\mathrm{part}}$, we verify three conditions against the predicted and ground-truth object boxes: (i) $\hat{b}^{\mathrm{part}}$ is spatially contained within at least one object box, (ii) it is not identical to that object box, and (iii) its area is strictly smaller, $|\hat{b}^{\mathrm{part}}| < |\hat{b}^{\mathrm{obj}}|$. Conditions (ii) and (iii) prevent the policy from trivially replicating the object box as the part prediction to satisfy containment without performing real part localization. The reward is the fraction of predicted parts satisfying all three conditions.

\paragraph{\textbf{Reflective Refinement Rewards.}}
These components incentivize the self-reflective loop in OP-HRG.

The \textit{improvement reward} encourages meaningful refinement from the initial to the final prediction. For IoU, it is defined as
\begin{equation}
R_{\mathrm{improv}}^{\mathrm{IoU}} = \max\!\Bigl(0,\; R_{\mathrm{IoU}}^{\mathrm{final}} - \max\bigl(R_{\mathrm{IoU}}^{\mathrm{initial}},\; \lambda_{\mathrm{IoU}} \cdot \mathrm{IoU}_{\mathrm{baseline}}\bigr)\Bigr),
\end{equation}
where $\mathrm{IoU}_{\mathrm{baseline}}$ is the precomputed IoU of a strong external baseline, so the model earns reward only when it surpasses the stronger of its own first answer or that baseline. This is critical for preventing reward exploitation: by rewarding the initial prediction independently and competing against the baseline, the policy has no incentive to produce deliberately poor first answers to inflate the improvement margin. We validate this design in Appendix~\ref{app:reward} (Fig.~\ref{fig:reward_exploit}), where removing the baseline reference causes the initial IoU to collapse toward zero. We similarly compute improvement rewards for the L1 and point rewards, and apply an explicit penalty $R_{\mathrm{IoU}}^{\mathrm{final}} - R_{\mathrm{IoU}}^{\mathrm{initial}}$ when the final IoU degrades relative to the initial prediction, discouraging harmful refinement.

The \textit{adjustment consistency penalty} penalizes contradictions between the model's adjustment intent and its actual behavior: declaring \texttt{ADJUSTMENT: YES} with identical initial and final coordinates, or \texttt{ADJUSTMENT: NO} with changed coordinates, incurs a penalty of $-1$, while consistent behavior incurs none. This enforces honest self-reflection and prevents gaming of the critique mechanism.

\subsection{Overall Training Objective}
\label{subsec:training_objective}
The pretrained multimodal LLM is adapted using GRPO with the composite reward above. Let $\pi_{\theta_0}$ denote the frozen reference policy and $\pi_\theta$ the updated policy. For each image--query pair $(I,q)$, GRPO samples a group $\mathcal{G}=\{y_j\}_{j=1}^{|\mathcal{G}|}$ of structured outputs, scores each with the normalized reward $R(y_j) = \sum_{k} \lambda_k R_k(y_j)$, and computes advantages $A_j$ as deviations from the group mean, eliminating the need for an explicit value function. The policy is optimized via a clipped surrogate objective with a KL-divergence regularizer:
\begin{equation}
L_{\mathrm{GRPO}} = -\frac{1}{|\mathcal{G}|}\sum_{j=1}^{|\mathcal{G}|}
      \min\!\left(
      \rho_j A_j,\,
      \mathrm{clip}(\rho_j,1\!-\!\epsilon,1\!+\!\epsilon)\,A_j
      \right)
      + \beta\,D_{\mathrm{KL}}\!\left(\pi_\theta\|\pi_{\theta_0}\right),
\end{equation}
where $\rho_j = \pi_\theta(y_j|I,q)\,/\,\pi_{\theta_0}(y_j|I,q)$ is the importance ratio, $\epsilon$ the clipping threshold, and $\beta$ the KL strength. The clipped surrogate restricts update magnitudes for training stability, while the KL term keeps the policy from drifting from its pretrained behavior, so that improvements in part-centric localization preserve object-level performance rather than degrading the model's broader competence.
\section{Experiments}
\label{sec:experiment}

\subsection{Models}

\textbf{Reasoning Model.} We use Qwen3-VL-Instruct as our reasoning component. We adopt the Instruct variant for two reasons. First, OP-HRG relies on explicit, step-by-step adherence to a structured prompt rather than free-form latent reasoning, and the instruction-tuned variant is well suited to following this fixed sequence of tagged steps. Second, on Qwen3-VL's own grounding evaluations~\cite{bai2025qwen3}, the Instruct variants attain stronger object detection performance than the Think variants at both the 4B and 8B scales, indicating they are the stronger starting point for spatial localization. We adopt the compact 4B model as our main configuration to demonstrate that structured reasoning and targeted rewards, rather than scale alone, drive the gains, and additionally report an 8B configuration (Table~\ref{tab:active_perception}) to show that the framework scales with backbone size. We do not perform a supervised fine-tuning cold-start, since Qwen3-VL is already instruction-tuned to emit bounding boxes and representative points. We apply the OP-HRG prompt structure and train from the pretrained weights with GRPO under our verifiable part-aware rewards, which directly reinforce and sharpen this existing capability rather than learning it from scratch.

\textbf{Mask Decoder.} For pixel-level segmentation, we employ the pretrained SAM2-Large model as our main frozen mask decoder. We use SAM2 for two reasons. First, prior baselines like \cite{visionreasoner} use the same decoder, and this ensures a fair comparison. Second, prior literature demonstrates that SAM-style models already produce high-quality masks from spatial prompts such as boxes and points. We therefore keep the SAM2 parameters frozen, isolating all improvements to the reasoning-guided localization abilities of the MLLM.

\subsection{Datasets and Preprocessing}
\label{sec:datasets}
\textbf{General Object Data.} 
We use the 7k multi-object dataset of \cite{visionreasoner}, which provides bounding boxes and representative points paired with free-form text queries. The purpose of this dataset is to train the model on visual grounding on a general, multi-object dataset that is not specific to object parts. 

For part-level supervision, we employ the InstructPart train split \cite{instructpart} containing 1200 images. The dataset provides curated, part-focused images with object–part structure. The dataset provides part masks but no bounding boxes or representative points. Since we need bounding boxes and points as localization primitives to train our reasoning LLM, we derive these from the masks: each connected component is converted into a bounding box, and the deepest interior pixel is selected via the Euclidean distance transform as the representative point.


\textbf{Baseline Signals.} As described in our reward design, the improvement reward credits gains only when the final prediction surpasses the stronger of the model's own initial answer and an external baseline. For this external reference, we use SAM3 \cite{sam3}, which generates segmentation masks from short textual phrases. We pre-compute baseline IoU scores by passing all training queries to SAM3, measuring predicted box overlap with ground truth, and storing the results as an additional field in the training data.

\subsection{Results}

\begin{table}[t]
\centering
\caption{Results on part-grounding benchmarks (gIoU). Our 4B model surpasses larger 7B grounding LLMs and SAM3 on cross-dataset zero-shot part grounding (PascalPart, PartImageNet), while leading on objects. $\dagger$InstructPart is evaluated on its test split; our RL training uses the InstructPart train split.}
\vspace{-6pt}
\label{tab:zero_shot_part}
\small
\resizebox{\linewidth}{!}{
\begin{tabular}{lcccc}
\toprule
& \multicolumn{3}{c}{\textbf{Parts}} & \multicolumn{1}{c}{\textbf{Objects}} \\
\cmidrule(lr){2-4} \cmidrule(lr){5-5}
\textbf{Model} & \textbf{InstructPart}$^\dagger$ & \textbf{PascalPart} & \textbf{PartImgNet} & \textbf{Pascal-Obj} \\
\midrule
\rowcolor{lightgray!55}{\textit{Token-based grounding MLLMs}} & & & & \\
LISA-7B \cite{lisa} & 43.26 & 13.82 & 29.91 & 83.50 \\
Sa2VA-4B \cite{yuan2025sa2va} & 50.15 & 14.67 & 38.13 & 77.02 \\
PixelLM-7B \cite{ren2024pixellm} & 44.41 & 16.57 & 35.25 & 81.71 \\
UniPixel-3B \cite{liu2026unipixel} & 59.68 & 30.64 & 46.18 & 85.54 \\
\midrule
\rowcolor{lightgray!55}{\textit{Decoupled MLLM-to-SAM}} & & & & \\
Molmo  + SAM3 (point) \cite{deitke2024molmo, sam3} & 51.02 & 8.87 & 17.30 & 57.57 \\
Grounding DINO + SAM3 (box) \cite{liu2024grounding, sam3} & 33.74 & 13.13 & 31.38 & 79.25 \\
VisionReasoner-7B \cite{visionreasoner} & 59.38 & 27.44 & 45.59 & 86.68 \\
\midrule
\rowcolor{lightgray!55}{\textit{Text-promptable segmentation}} & & & & \\
ClipSeg \cite{luddecke2022image} & 31.85 & 12.51 & 33.45 &  70.51 \\
SAM3 (text) \cite{sam3} & 71.06 & 33.05 & 53.89 & 85.32 \\
\midrule
\rowcolor{lightgray!55} \multicolumn{5}{l}{{\textit{Qwen3-VL-Instruct-4B + SAM2}}} \\
\quad zero-shot OP-HRG prompt & 31.45 & 12.83 & 21.95 & 36.91 \\
\rowcolor{blue!15} \quad + RL using InstructPart \textbf{(Ours)} & \cellcolor{blue!15} \textbf{75.56} & \cellcolor{blue!15} \textbf{38.59} & \cellcolor{blue!15} \textbf{56.87} & \cellcolor{blue!15} \textbf{87.50} \\
\midrule
\textit{$\Delta$ vs.\ best baseline} & \textit{+4.50} & \textit{+5.54} & \textit{+2.98} & \textit{+0.82} \\
\bottomrule
\end{tabular}
}
\end{table}

We evaluate on three benchmarks: InstructPart \cite{instructpart} (600 queries, 600 images), PascalPart \cite{chen2014detect} (specifically the PascalPart-116 split \cite{wei2023ov}, which contains 1K object + 10K part queries across 851 images), and PartImageNet \cite{he2022partimagenet} (14K part queries across 4589 images), reporting gIoU (the mean IoU across all test queries). InstructPart is evaluated on the test split of the dataset used for part-level training, whereas PascalPart and PartImageNet are fully cross-dataset zero-shot -- our model has seen neither images nor annotations from either benchmark. Both feature diverse, naturally occurring scenes at far larger scale than the InstructPart training set, making them a challenging test of generalization. Therefore, the gains there show that training on just 1200 part-focused images with our method improves part grounding.

Table \ref{tab:zero_shot_part} compares our pipeline against baselines that span different paradigms for visual grounding. We first compare against a set of special-token grounding LLMs (LISA-7B \cite{lisa}, Sa2VA-4B \cite{yuan2025sa2va}, PixelLM-7B \cite{ren2024pixellm}, and UniPixel-3B \cite{liu2026unipixel}), which embed segmentation tokens into the LLM output to condition a jointly trained mask decoder. All underperform our method on parts, with the strongest (UniPixel-3B) trailing by 15.88, 7.95, and 10.69 gIoU on InstructPart, PascalPart, and PartImageNet parts.

Next, we compare against decoupled MLLM-to-SAM pipelines, which, like ours, prompt a frozen mask decoder with MLLM predictions. Since this design relies on the MLLM producing bounding boxes and representative points as prompts, we first consider two baselines that isolate each modality. Molmo \cite{deitke2024molmo}, a strong pointing model, is paired with SAM3 to generate masks from single-point predictions; Grounding DINO \cite{liu2024grounding}, a state-of-the-art open-set detector, is paired with SAM3 to generate masks from predicted bounding boxes. Both perform substantially worse than our method on parts, confirming that neither strong pointing nor strong detection alone suffices for part grounding. VisionReasoner \cite{visionreasoner} pairs a Qwen2.5-VL-7B model with SAM2 and uses RL alignment for grounding; with a smaller 4B model, our method improves on it by +16.18, +11.15, and +11.28 gIoU on InstructPart, PascalPart, and PartImageNet parts, demonstrating the effectiveness of structured hierarchical reasoning and part-aware rewards. SAM3 \cite{sam3}, the latest model in the Segment Anything family, is trained to segment visual concepts from textual phrases and is the strongest baseline. Nevertheless, our method outperforms SAM3 both on in-domain (InstructPart) and zero-shot parts evaluation. Finally, we compare against our base model: Qwen3-VL-Instruct-4B paired with SAM2 under the OP-HRG prompt, without any RL training. This vanilla configuration scores far below our trained model across every benchmark, on the identical architecture and prompt, isolating the impact of our reinforcement learning framework with part-aware rewards. Beyond parts, our method also achieves the strongest object-level performance (87.50 gIoU on Pascal-Obj) compared to all baselines in Table \ref{tab:zero_shot_part}. This shows that part-centric training improves object-level grounding on these benchmarks.

\begin{table*}[t]
\centering

\begin{minipage}[t]{0.46\linewidth}
\vspace{0pt}

\captionof{table}{Frozen mask-decoder swap with our trained MLLM fixed (gIoU).}
\vspace{-10pt}
\label{tab:decoder_swap}
\setlength{\tabcolsep}{4pt}
\footnotesize
\resizebox{\linewidth}{!}{%
\begin{tabular}{lcccc}
\toprule
& \multicolumn{3}{c}{\textbf{Parts}} & \textbf{Obj} \\
\cmidrule(lr){2-4} \cmidrule(lr){5-5}
\textbf{Decoder} & \textbf{InstP}$^\dagger$ & \textbf{PascP} & \textbf{PartIN} & \textbf{PascO} \\
\midrule
\rowcolor{blue!15} SAM2 \textbf{(Ours)} & 75.56 & 38.59 & 56.87 & 87.50 \\
SAM3 & 75.77 & 39.05 & 56.23 & 87.73 \\
\bottomrule
\end{tabular}}

\vspace{6pt}  

\captionof{table}{Active visual perception.}
\vspace{-10pt}
\label{tab:active_perception}
\setlength{\tabcolsep}{4pt}
\footnotesize
\resizebox{\linewidth}{!}{%
\begin{tabular}{lcccc}
\toprule
& \multicolumn{3}{c}{\textbf{Parts}} & \textbf{Obj} \\
\cmidrule(lr){2-4} \cmidrule(lr){5-5}
\textbf{Model} & \textbf{InstP}$^\dagger$ & \textbf{PascP} & \textbf{PartIN} & \textbf{PascO} \\
\midrule
Ours (4B) & 75.56 & 38.59 & 56.87 & 87.50 \\
\;+ AP (4B) & 76.34 & 39.04 & 55.98 & 88.09 \\
\rowcolor{blue!15} AP (8B) & \textbf{77.00} & \textbf{40.28} & \textbf{58.19} & \textbf{89.62} \\
\bottomrule
\end{tabular}}
\end{minipage}
\hfill
\begin{minipage}[t]{0.50\linewidth}
\vspace{0pt} 

\captionof{table}{Referring grounding (Acc@0.5) on RefCOCO/+/g.}
\vspace{-10pt}
\label{tab:refcoco}
\centering
\setlength{\tabcolsep}{4pt}
\footnotesize
\resizebox{\linewidth}{!}{%
\begin{tabular}{lcccc}
\toprule
\textbf{Model} & \textbf{RefC} & \textbf{RefC+} & \textbf{RefCg} & \textbf{Avg} \\
 & \textbf{testA} & \textbf{testA} & \textbf{test} & \\
\midrule
\rowcolor{lightgray!55} \multicolumn{5}{l}{{\textit{Specialist grounding models}}} \\
GLIP-T (ZS) \cite{li2022grounded} & 54.3 & 52.8 & 66.9 & 58.0 \\
MDETR \cite{kamath2021mdetr} & 89.6 & 84.1 & 80.9 & 84.9 \\
GDINO-T \cite{liu2024grounding} & 91.9 & 87.4 & 84.9 & 88.1 \\
\midrule
\rowcolor{lightgray!55} \multicolumn{5}{l}{{\textit{Multimodal LLMs}}} \\
Shikra-7B \cite{chen2023shikra} & 90.6 & 87.4 & 82.2 & 86.7 \\
InternVL2-8B \cite{chen2024internvl} & 91.1 & 87.9 & 82.7 & 87.2 \\
Qwen2.5-VL-7B \cite{bai2025qwen25vl} & 91.7 & 88.2 & 85.7 & 88.5 \\
Qwen3-VL-Instruct-4B \cite{bai2025qwen3} & 92.7 & 88.7 & 87.8 & 89.7 \\
VisionReasoner-7B \cite{visionreasoner} & 90.6 & 87.9 & 87.5 & 88.7 \\
\rowcolor{blue!15} Ours-4B & 89.3 & 83.3 & 86.9 & 86.5 \\
\bottomrule
\end{tabular}}
\end{minipage}
\end{table*}

Our decoupled design treats the mask decoder as a frozen, interchangeable module, raising the question of whether our gains depend on the specific decoder rather than the reasoning-guided localization. To test this, we hold our trained Qwen3-VL-Instruct-4B fixed and replace only the decoder, prompting SAM3 \cite{sam3} with the same predicted boxes and points in place of SAM2. As Table \ref{tab:decoder_swap} shows, the swap does not significantly shift the gIoU, confirming that the quality of our MLLM-generated prompts, not the decoder, drives performance; we retain SAM2 by default to match prior decoupled pipelines \cite{visionreasoner}.

Table~\ref{tab:active_perception} reports our model with the active visual perception step introduced in Section \ref{sec:active_perception}. At 4B, conditioning the final answer on re-encoded crops improves performance on three of the four splits. The 8B configuration improves over its 4B counterpart, demonstrating that our framework scales to larger models. 

We next verify that part-centric training does not compromise general object-level grounding. Table \ref{tab:refcoco} reports referring grounding accuracy (Acc@0.5) on RefCOCO/+/g. Our 4B model attains 86.5 average accuracy, competitive with the 7B VisionReasoner~\cite{visionreasoner} (88.7) and within roughly three points of its own Qwen3-VL-Instruct-4B base~\cite{bai2025qwen3} (89.7). This trade-off is modest relative to the large part-grounding gains in Table \ref{tab:zero_shot_part}, indicating that reinforcing fine-grained part localization largely preserves whole-object referring. Beyond explicit part and object grounding, we assess whether our structured reasoning transfers to tasks requiring implicit multi-step inference. Table \ref{tab:reasonseg} reports gIoU on ReasonSeg, a reasoning-driven segmentation benchmark. Our 4B model reaches 69.6 gIoU, surpassing the baselines and indicating that the reasoning induced by OP-HRG benefits segmentation broadly, not only part localization.

Finally, we examine inference cost. Table \ref{tab:inference_cost} reports MLLM wall-clock time and average generated tokens on an NVIDIA L40 at batch size 32. Our method runs faster than the same base model under the OP-HRG prompt without RL, because GRPO training discourages unconstrained reasoning and yields more compact outputs (564 vs.\ 607 tokens). Relative to VisionReasoner, our hierarchical, self-reflective procedure adds about one second of overhead, a modest cost given the substantial part-grounding gains in Table \ref{tab:zero_shot_part}. We provide qualitative examples illustrating our model's predictions across all benchmarks in Appendix~\ref{app:qualitative}.

\begin{table*}[t]
\centering
\begin{minipage}[t]{0.39\linewidth}
\centering
\vspace{0pt} 

\captionof{table}{Reasoning segmentation on ReasonSeg (gIoU). Our 4B model surpasses baselines, including the 7B VisionReasoner, showing that part-centric training also benefits reasoning-driven object segmentation.}
\label{tab:reasonseg}
\setlength{\tabcolsep}{5pt}
\scriptsize
\begin{tabular}{lc}
\toprule
\textbf{Model} & \textbf{ReasonSeg} \\
\midrule
ReLA \cite{liu2023gres} & 21.3 \\
LISA-7B \cite{lisa} & 36.8 \\
Qwen2.5-VL-7B \cite{bai2025qwen25vl} & 52.1 \\
SegZero-7B \cite{segzero} & 57.5 \\
VisionReasoner-7B \cite{visionreasoner} & 63.6 \\
GenSeg-R1-4B \cite{hegde2026genseg} & 68.4 \\
\rowcolor{blue!15} \textbf{Ours-4B} & \textbf{69.6} \\
\bottomrule
\end{tabular}
\end{minipage}
\hfill
\begin{minipage}[t]{0.56\linewidth}
\centering
\vspace{0pt} 

\captionof{table}{Inference cost comparison}
\vspace{-5pt}
\label{tab:inference_cost}
\setlength{\tabcolsep}{6pt}
\scriptsize
\begin{tabular}{lcc}
\toprule
\textbf{Model} & \textbf{Time (s)} & \textbf{Avg.\ tokens} \\
\midrule
VisionReasoner-7B~\cite{visionreasoner} & 1.86 & 272 \\
ZS OP-HRG Prompt & 3.31 & 607 \\
\rowcolor{blue!15} Ours & 2.87 & 564 \\
\bottomrule
\end{tabular}

\vspace{11pt} 

\captionof{table}{Ablation on the InstructPart test set.}
\vspace{-5pt}
\label{tab:ablation_instructpart}
\setlength{\tabcolsep}{7pt}
\scriptsize
\begin{tabular}{lc}
\toprule
\textbf{Model} & \textbf{gIoU} \\
\midrule
\rowcolor{lightgray!55} \multicolumn{2}{l}{{\textit{(a) Is part-level training data sufficient?}}} \\
VisionReasoner (Qwen2.5-VL-7B + SAM2) & 59.38 \\
\quad + GRPO on InstructPart & 62.33 \\
\midrule
\rowcolor{lightgray!55} \multicolumn{2}{l}{{\textit{(b) Are both OP-HRG components necessary?}}} \\
Ours w/o hierarchy \& refinement & 70.32 \\
Ours w/o hierarchical grounding & 73.77 \\
Ours w/o reflective refinement & 73.98 \\
\rowcolor{blue!15} \textbf{Ours} & \cellcolor{blue!15}\textbf{75.56} \\
\bottomrule
\end{tabular}
\end{minipage}
\end{table*}

\subsection{Ablation Studies}

\begin{figure}[t]
  \centering
  \includegraphics[width=0.98\linewidth]{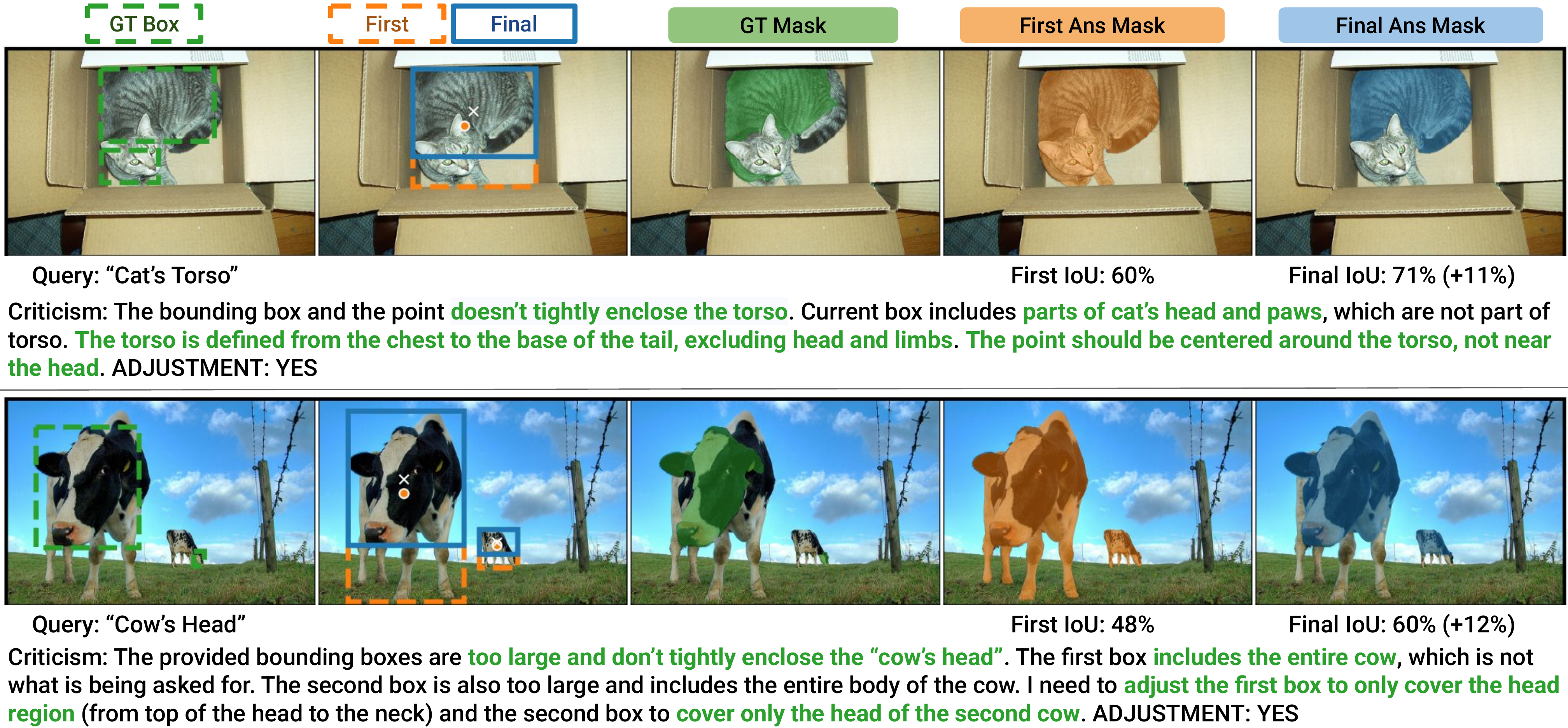}
  
  
  \caption{Examples of the reflective step from an intermediate model checkpoint.}
  \label{fig:refinement_examples}
\end{figure}

We conduct ablation experiments on the InstructPart test set~\cite{instructpart}, as shown in Table~\ref{tab:ablation_instructpart}. We choose InstructPart for this analysis because it is drawn from the same domain as our part-level training data, allowing us to directly isolate the contribution of each architectural and reward component without conflating the effects of cross-dataset distribution shift.

\textbf{Is part-level training data sufficient?}
VisionReasoner~\cite{visionreasoner} achieves 59.38 gIoU on InstructPart using its original training and reward design. To test whether simply exposing a strong baseline to part data closes the gap, we retrain VisionReasoner on the InstructPart train split using GRPO with its original reward function, without OP-HRG prompting or our part-aware rewards. This yields a gain to 62.33 gIoU vs our 75.56, indicating that part data adaptation alone is insufficient.

\textbf{Are both object-part hierarchy and the reflective refinement components necessary?} We ablate the two core mechanisms of OP-HRG independently. Training with only standard localization rewards (IoU, L1, point accuracy) and no OP-HRG structure reaches 70.32 gIoU, which confirms that GRPO alignment helps but trails our full method. Removing hierarchical grounding -- the object-first localization stage, the part containment constraint, and the object hint reward -- while keeping reflective refinement reduces performance to 73.77 gIoU; conversely, removing reflective refinement -- the self-critique loop, the two-answer structure, and the improvement reward -- while keeping hierarchical grounding gives 73.98 gIoU. Both underperform the full pipeline, showing that hierarchical grounding and reflective refinement are complementary and individually necessary.

\subsection{Analyzing the Reflective Step} 
\label{sec:causal_refinement}

The ablation above shows that reflective refinement contributes a modest gain (+1.58 gIoU on InstructPart). We find that this gain acts primarily as a train-time regularizer rather than a test-time process: a model trained with the full objective but evaluated with a plain single-answer prompt retains nearly all of it (75.40 vs.\ 75.56), so the benefit is internalized into the weights rather than supplied by the prompt at test time.

\begin{wrapfigure}{r}{0.5\linewidth}
  \centering
  \vspace{-25pt}
  \includegraphics[width=\linewidth]{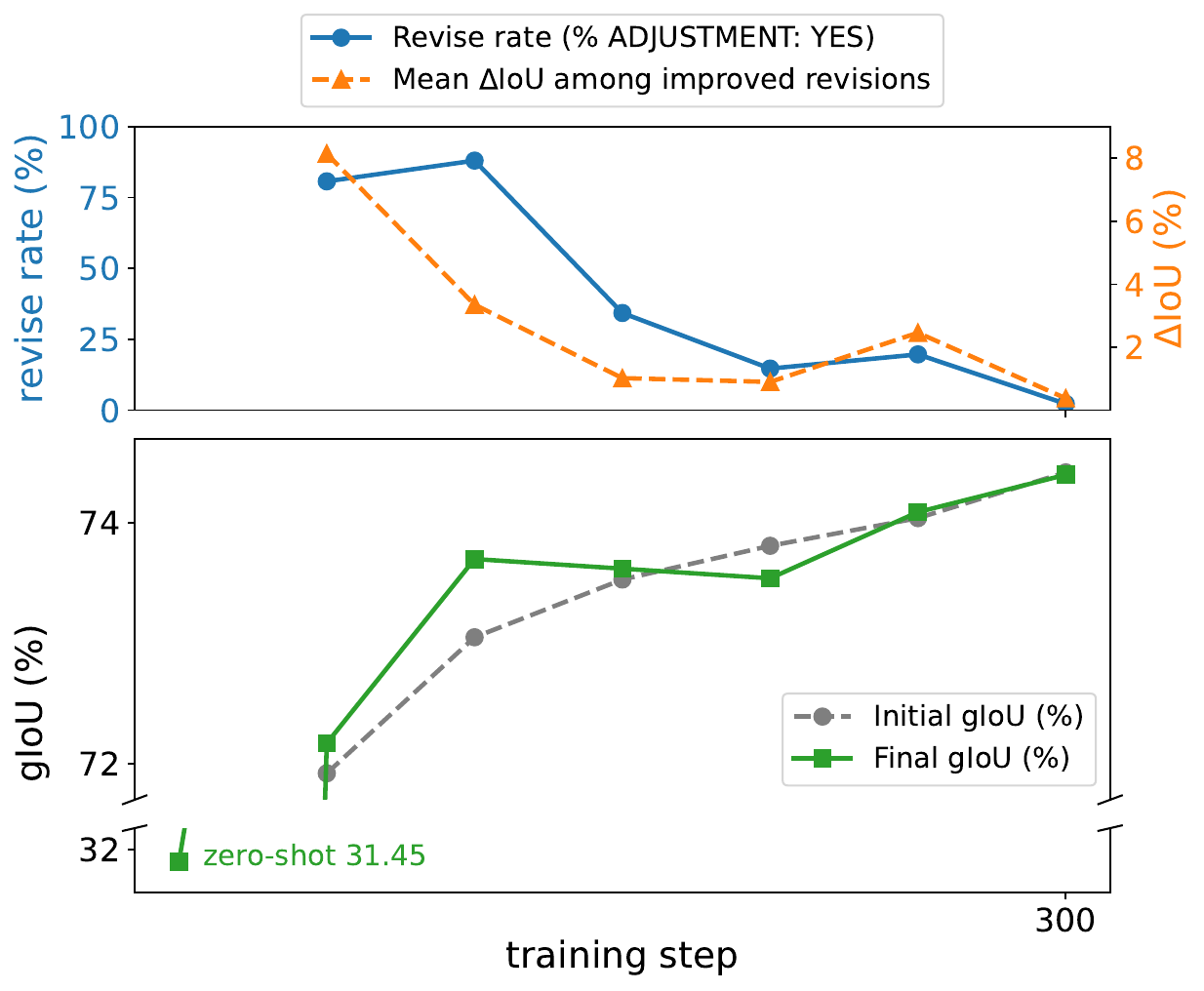}
  \caption{Refinement behavior over training. Top: revision rate and mean IoU gain. Bottom: initial vs.\ final gIoU.}
  \vspace{-23pt}
  \label{fig:refine_trajectory}
\end{wrapfigure}

\textbf{Refinement is active early and progressively internalized.} We checkpoint the model every 50 steps and trace its refinement behavior on InstructPart (Figure \ref{fig:refine_trajectory}). Early in training the model revises often, with a net positive IoU change (top), confirming the reflective step does real corrective work. Figure \ref{fig:refinement_examples} shows two such cases. The model tightens an over-broad box that had included the cat's head and paws down to just the torso (IoU 0.60 vs.\ 0.71), and narrows a whole-cow box to the queried cow's head alone (0.48 vs.\ 0.60). The bottom panel in Fig. \ref{fig:refine_trajectory} shows the initial- and final-answer gIoU: early on, the final sits above the initial, but the two converge as training proceeds. Once the model achieves its best possible first answer, the reflective step increasingly acts as verification rather than correction, as further mentioned in Sec. \ref{sec:limitations}.

\section{Limitations}
\label{sec:limitations}

We observe that as training progresses over many steps, the reasoning model converges toward producing higher quality predictions at the initial stage, which in turn leads to the reflective mechanism consistently declining to adjust. This is a natural consequence of optimization: as the model's first-answer accuracy improves towards the best that it can achieve, there is less need for refinement, and our reward discourages unnecessary changes. However, this convergence effectively reduces the reflective refinement loop to a verification step rather than an active correction mechanism. Future work could explore the utility of the reflective stage even as the base localization quality improves. Our part containment reward is satisfied whenever a predicted part box falls within \textit{any} parent object box. A part localized within the wrong object of the same category can still be scored as valid. As most of our InstructPart training data does not contain such multi-instance scenes, this has limited effect, and we leave this to future work.

\section{Conclusion}

We presented Object-Part Hierarchical Reflective Grounding (OP-HRG), which addresses a basic weakness of current multimodal LLMs that ground whole objects well but fail on fine-grained parts. OP-HRG works coarse-to-fine. It first localizes the parent object, then the part within it, and refines and verifies the result with a self-check. We train it with a part-aware GRPO framework whose stage-wise rewards supervise each step. With a compact 4B-parameter model, our method outperforms larger baselines on PascalPart and PartImageNet (cross-dataset zero-shot) and InstructPart (in-domain) with only a modest trade-off in general object referring, and it transfers to reasoning segmentation. Ablations show that hierarchical grounding and reflective refinement contribute comparable, complementary gains, with the reflective step's benefit largely internalized during training. These results suggest that the capacity for fine-grained spatial reasoning already exists within pretrained MLLMs and can be drawn out through structured prompting and targeted reward design. 
\section{Acknowledgments}

This research is supported by grants from the National Science Foundation (NSF) for the HDR Imageomics Institute
(OAC-2118240). We are thankful for the computational resources provided by the Advanced Research
Computing Center (ARC) at Virginia Tech and the Ohio Supercomputer Center.

\bibliographystyle{splncs04}
\bibliography{main}

\newpage
\appendix
\section{Overview of the Appendix}

This supplementary material provides additional details that complement the main paper. Section~\ref{app:prompt} presents the full OP-HRG training and inference prompt. Section~\ref{app:reward} gives the complete implementation of our reward function, including hyperparameters, component weights, normalization, and an empirical validation of the improvement reward design. Section~\ref{app:verl} details our VeRL-based \cite{sheng2024hybridflow} training framework, including the technical challenges of extending it for active perception (Section~\ref{app:ap_challenges}). Section~\ref{app:eval} describes our evaluation protocol, and Section~\ref{app:extended_results} reports extended gIoU and cIoU results on the part-grounding benchmarks. Section~\ref{app:qualitative} presents qualitative segmentation examples and reasoning traces across our evaluation benchmarks. Section~\ref{app:more_ablations} reports additional ablations on backbone selection and the contribution of the structured prompt (Sections~\ref{sec:instruct_vs_think} and~\ref{sec:bare_prompt}), and Section~\ref{app:provenance} documents the provenance of all baseline numbers reported in the paper.

\section{Prompts}
\label{app:prompt}

\begin{promptbox}[title=Full OP-HRG Prompt]
Please find "{Question}" with bounding boxes and points from the given image.
Your goal is to output tight bounding boxes and a representative point.

GENERAL RULES:
- Output tight, compact boxes with minimal extra background.
- The representative point must lie inside "{Question}".
- If nothing matches the query, output [] for the relevant JSON lists.
- The query may refer to a whole object or a part of an object. If the query is
  about a part, first find the whole object that contains the part, then find the
  part within that object.

You MUST follow these steps and output structure.

STEP 1 - Locate
In <locate> </locate> tags, think about:
- What the query is asking for.
- Whether it is a whole object or a part.
- Where it is in the image.

STEP 2 - Decide if you are finding an object or a part
In <target> </target> tags, output either "object" or "part":
- "object": the query is about a whole object.
- "part": the query is about a part of an object.

STEP 3 - If <target> is a "part", locate the object to which the part belongs
- If <target> is "part", first find the whole object(s) that contain the part.
  In <object_hint> </object_hint> tags, output a JSON list of these object boxes:
  [{"bbox_2d": [x1, y1, x2, y2], "point_2d": [cx, cy]}, ...]
- If <target> is "object", inside <object_hint> tags output a single empty list: [].

STEP 4 - First Answer
In <first_answer> </first_answer> tags, output a JSON list of the boxes for what
the query directly refers to, that is "{Question}":
  [{"bbox_2d": [px1, py1, px2, py2], "point_2d": [pcx, pcy]}, ...]
- If <target> is "object", these boxes should cover the whole object.
- If <target> is "part", these boxes should cover only the part (not the entire
  object). The part should be within the object boxes in <object_hint>.

STEP 5 - Criticism (Self-Check)
In <criticism> </criticism> tags, re-examine your <first_answer>. Do the boxes
tightly enclose "{Question}"? If adjustments are needed, describe the issue and
suggested adjustments. Examples of necessary adjustments could be to make the
bboxes smaller or bigger, or move the bboxes in any direction.
The last part inside <criticism> MUST be exactly one of:
- "ADJUSTMENT: YES"
- or "ADJUSTMENT: NO"
Output "ADJUSTMENT: NO" if the boxes and points in <first_answer> are correct and
no adjustments are necessary. Output "ADJUSTMENT: YES" if the <first_answer> needs
adjustments to enclose "{Question}".
Format (describe what YOU actually observe; do not copy this wording):
- "<criticism>[your assessment of whether the boxes tightly enclose the target,
  and the specific adjustment you will make, if any]. ADJUSTMENT: YES|NO</criticism>"

STEP 6 - FINAL ANSWER
In <answer> </answer> tags, output the final answer.
- If <criticism> ends with "ADJUSTMENT: NO":
  - The JSON list in <answer> MUST be IDENTICAL to the JSON list in <first_answer>.
- If <criticism> ends with "ADJUSTMENT: YES":
  - At least one bbox_2d or point_2d in <answer> MUST be different from those in
    <first_answer>, and the changes should address the issues described in <criticism>.
The <answer> should contain a JSON list of entries with bbox_2d and point_2d:
  [{"bbox_2d": [qx1, qy1, qx2, qy2], "point_2d": [qcx, qcy]}, ...]

OUTPUT FORMAT EXAMPLE (STRUCTURE ONLY):
<locate>thinking here</locate>
<target>object|part</target>
<object_hint>[{"bbox_2d": [x1,y1,x2,y2], "point_2d": [cx,cy]}, ...] | []</object_hint>
<first_answer>[{"bbox_2d": [px1,py1,px2,py2], "point_2d": [pcx,pcy]}, ...]</first_answer>
<criticism>criticism here. ADJUSTMENT: YES|NO</criticism>
<answer>[{"bbox_2d": [qx1,qy1,qx2,qy2], "point_2d": [qcx,qcy]}, ...]</answer>
\end{promptbox}

\begin{promptbox}[title=Active Perception Input Prompt]
Please find "{Question}" with bounding boxes and points from the given image.
Your goal is to output tight bounding boxes and a representative point.

GENERAL RULES:
- Output tight, compact boxes with minimal extra background.
- The representative point must lie inside "{Question}".
- If nothing matches the query, output [] for the relevant JSON lists.
- The query may refer to a whole object or a part of an object. If the query is
  about a part, first find the whole object that contains the part, then find the
  part within that object.

You MUST follow these steps and output structure.
After completing Step 4 you MUST stop. Do NOT continue beyond </first_answer>.

STEP 1 - Locate
In <locate> </locate> tags, think about:
- What the query is asking for.
- Whether it is a whole object or a part.
- Where it is in the image.

STEP 2 - Decide if you are finding an object or a part
In <target> </target> tags, output either "object" or "part":
- "object": the query is about a whole object.
- "part": the query is about a part of an object.

STEP 3 - If <target> is a "part", locate the object to which the part belongs
- If <target> is "part", first find the whole object(s) that contain the part.
  In <object_hint> </object_hint> tags, output a JSON list of these object boxes:
  [{"bbox_2d": [x1, y1, x2, y2], "point_2d": [cx, cy]}, ...]
- If <target> is "object", inside <object_hint> tags output a single empty list: [].

STEP 4 - First Answer
In <first_answer> </first_answer> tags, output a JSON list of the boxes for what
the query directly refers to, that is "{Question}":
  [{"bbox_2d": [px1, py1, px2, py2], "point_2d": [pcx, pcy]}, ...]
- If <target> is "object", these boxes should cover the whole object.
- If <target> is "part", these boxes should cover only the part (not the entire
  object). The part should be within the object boxes in <object_hint>.

STOP HERE. The user will provide crops of your predicted regions for you
to review before you continue.

OUTPUT FORMAT EXAMPLE (STRUCTURE ONLY):
<locate>thinking here</locate>
<target>object|part</target>
<object_hint>[{"bbox_2d": [x1,y1,x2,y2], "point_2d": [cx,cy]}, ...] | []</object_hint>
<first_answer>[{"bbox_2d": [px1,py1,px2,py2], "point_2d": [pcx,pcy]}, ...]</first_answer>
\end{promptbox}

\begin{promptbox}[title=Active Perception -- Injected Crop Turn]
<|im_end|>
<|im_start|>user
<crop>
Here are crops of your predicted bounding box regions from the
original image:
Region 1 (original image coordinates: [x1, y1, x2, y2]):
<|vision_start|><|image_pad|><|image_pad|>...<|vision_end|>
Region 2 (original image coordinates: [x1, y1, x2, y2]):
<|vision_start|><|image_pad|><|image_pad|>...<|vision_end|>
</crop>

IMPORTANT: The cropped images above are extra context only.
Always propose bounding boxes using coordinates from the ORIGINAL image,
not relative to the crops.

Now continue with Step 5 and Step 6.

STEP 5 - Criticism (Self-Check)
In <criticism> </criticism> tags, examine the crops above and check
your <first_answer>. Do the boxes tightly enclose "{query}"? If adjustments are
needed, describe the issue and suggested adjustments. Examples of necessary
adjustments could be to make the bboxes smaller or bigger, or move the bboxes
in any direction.
The last part inside <criticism> MUST be exactly one of:
- "ADJUSTMENT: YES"
- or "ADJUSTMENT: NO"
Output "ADJUSTMENT: NO" if the boxes and points in <first_answer> are correct and
no adjustments are necessary. Output "ADJUSTMENT: YES" if the <first_answer> needs
adjustments to enclose "{query}".
Format (describe what YOU actually observe in the crops; do not copy this wording):
- "<criticism>[your assessment of whether the boxes tightly enclose the target,
  and the specific adjustment you will make, if any]. ADJUSTMENT: YES|NO</criticism>"

STEP 6 - FINAL ANSWER
In <answer> </answer> tags, output the final answer.
- If <criticism> ends with "ADJUSTMENT: NO":
  - The JSON list in <answer> MUST be IDENTICAL to the JSON list in <first_answer>.
- If <criticism> ends with "ADJUSTMENT: YES":
  - At least one bbox_2d or point_2d in <answer> MUST be different from those in
    <first_answer>, and the changes should address the issues described in <criticism>.
The <answer> should contain a JSON list of entries with bbox_2d and point_2d:
  [{"bbox_2d": [qx1, qy1, qx2, qy2], "point_2d": [qcx, qcy]}, ...]

OUTPUT FORMAT EXAMPLE (STRUCTURE ONLY):
<criticism>criticism here. ADJUSTMENT: YES|NO</criticism>
<answer>[{"bbox_2d": [qx1, qy1, qx2, qy2], "point_2d": [qcx, qcy]}, ...]</answer><|im_end|>
<|im_start|>assistant
\end{promptbox}

\section{Reward Function Implementation Details}
\label{app:reward}
This section provides complete implementation details of the reward function described in Section \ref{sec:reward_design}.

\subsection{Hyperparameters}

Table~\ref{tab:reward_hyperparams} lists all scalar hyperparameters used in the localization reward components.

\begin{table}[ht]
\centering
\caption{Reward function hyperparameters.}
\label{tab:reward_hyperparams}
\begin{tabular}{llp{7cm}}
\toprule
\textbf{Parameter} & \textbf{Value} & \textbf{Description} \\
\midrule
$\alpha_{\ell_1}$              & 0.10 & Scaling factor for adaptive L1 box coordinate threshold \\
$\tau^{(\ell_1)}_{\min}$       & 3    & Minimum cap on L1 threshold (pixels) \\
$\tau^{(\ell_1)}_{\max}$       & 10   & Maximum cap on L1 threshold (pixels) \\
$\alpha_{p}$                   & 0.20 & Scaling factor for adaptive point distance threshold \\
$\tau^{(p)}_{\min}$            & 5    & Minimum cap on point distance threshold (pixels) \\
$\tau^{(p)}_{\max}$            & 30   & Maximum cap on point distance threshold (pixels) \\
\bottomrule
\end{tabular}
\end{table}

The scaling factors $\alpha_{\ell_1}$ and $\alpha_p$ set how tightly predicted boxes and points must align with ground truth, while the clamps $\tau_{\min}$ and $\tau_{\max}$ bound the resulting thresholds in pixel space. Our choices are guided by a well-known tension in reward design for RL-based grounding: fixed, loose IoU-style thresholds are too permissive -- predictions that merely shift around the target can still receive near-maximal reward, producing reward ambiguity that degrades localization in later training -- while overly strict thresholds cause reward sparsity, where few sampled rollouts clear the bar, the intra-group reward variance collapses, and the GRPO advantage vanishes. Effective thresholds must therefore sit between these regimes, and the appropriate absolute tolerance depends on the size of the target region. We address this by scaling the box and point tolerances by the ground-truth box diagonal $d_j$, so that accuracy is judged relative to region size: a small part box is held to a proportionally tighter standard than a large object box, which a fixed pixel threshold would treat too leniently. The clamps prevent this adaptive tolerance from becoming degenerate for extremely small or large regions. We use a tighter scaling factor for box coordinates ($\alpha_{\ell_1}\!=\!0.10$) than for representative points ($\alpha_p\!=\!0.20$), since a point need only fall within the target region whereas box boundaries must align tightly with ground truth. These values can be tightened or loosened to make the localization rewards stricter or looser, however, we do not perform any per-benchmark tuning of these values.

\subsection{Component Weights and Reward Normalization}

All reward components are weighted equally at $1.0$, except the IoU reward, which is weighted at $2.0$ as the most direct measure of localization quality; this avoids any per-benchmark reward engineering. The total reward is then normalized by the maximum achievable score, which differs between object and part queries. For an \textbf{object query}:
\begin{equation}
R_{\max}^{\text{obj}} = \lambda_{\text{fmt}} + \lambda_{\text{dec}} + 2(\lambda_{\text{iou}} + \lambda_{\ell_1} + \lambda_{\text{pt}}) + \lambda_{\text{cmpct}} + \lambda_{\text{rep}},
\end{equation}
and for a \textbf{part query}:
\begin{equation}
R_{\max}^{\text{part}} = \lambda_{\text{fmt}} + \lambda_{\text{dec}} + \lambda_{\text{hint}} + 2(\lambda_{\text{iou}} + \lambda_{\ell_1} + \lambda_{\text{pt}}) + \lambda_{\text{cmpct}} + \lambda_{\text{contain}} + \lambda_{\text{rep}},
\end{equation}
where $\lambda_{\text{fmt}}$, $\lambda_{\text{dec}}$, $\lambda_{\text{hint}}$, $\lambda_{\text{iou}}$, $\lambda_{\ell_1}$, $\lambda_{\text{pt}}$, $\lambda_{\text{cmpct}}$, $\lambda_{\text{contain}}$, and $\lambda_{\text{rep}}$ are the weights for the format, decision, object hint, IoU, L1, point, compactness, part containment, and non-repetition rewards respectively. The factor of 2 on the localization terms reflects that IoU, L1, and point rewards are each applied twice: once for the initial prediction (\texttt{<first\_answer>}) and once for the final prediction (\texttt{<answer>}). The improvement reward and the adjustment consistency penalty are excluded from the normalization denominator: the improvement reward is bounded by the gap between the final and initial localization rewards and thus cannot exceed the localization reward weights, so including it would under-normalize the reward; the adjustment consistency reward is penalty-only with a maximum value of $0$, and therefore contributes nothing to the maximum achievable reward.

\subsection{Improvement Reward Details}

The improvement reward is computed separately for IoU, L1, and point components. For IoU, the reward is:

\begin{equation}
R_{\text{improv}}^{\text{IoU}} = \max\!\Bigl(0,\; R_{\text{IoU}}^{\text{final}} - \max\bigl(R_{\text{IoU}}^{\text{initial}},\; \lambda_{\text{iou}} \cdot \mathrm{IoU}_{\text{baseline}}\bigr)\Bigr),
\end{equation}

where $\lambda_{\text{iou}} = 2.0$ is the IoU reward weight and $\mathrm{IoU}_{\text{baseline}}$ is the precomputed baseline IoU. Note that scaling the baseline by $\lambda_{\text{iou}}$ places it on the same scale as $R_{\text{IoU}}^{\text{final}}$ and $R_{\text{IoU}}^{\text{initial}}$, which are already weighted. For L1 and point, the improvement reward is simply $\max(0, R^{\text{final}} - R^{\text{initial}})$, with no external baseline reference. An explicit drop penalty equal to $R_{\text{IoU}}^{\text{final}} - R_{\text{IoU}}^{\text{initial}}$ (a negative value) is added when the final IoU degrades relative to the initial prediction.

\subsection{Format Compliance Scoring}

The format reward combines a binary full-sequence match score with a content validity score. The binary score checks whether the complete output matches the expected tag sequence via a single regex. The content score assigns partial credit per tag: up to 1.0 for a well-formed \texttt{<object\_hint>} block (validated differently depending on whether the query is for a part or object), up to 1.0 for \texttt{<first\_answer>} (0.5 for a valid \texttt{bbox\_2d}, 0.5 for a valid \texttt{point\_2d}), up to 1.0 for \texttt{<answer>} (same breakdown), and up to 1.0 for a \texttt{<criticism>} block containing a valid \texttt{ADJUSTMENT: YES/NO} declaration. The binary score and content score are summed and divided by 5.0, yielding a normalized format reward in $[0, 1]$.

\subsection{External Baseline IoU}

The external baseline IoU ($\mathrm{IoU}_{\text{baseline}}$) is precomputed offline and stored alongside each training sample prior to training, avoiding any online inference overhead during GRPO training. When the baseline model produces no predictions for a given sample, $\mathrm{IoU}_{\text{baseline}}$ is set to $0.0$, in which case the improvement reward reduces to competing against the model's own initial prediction only.

\subsection{Empirical Validation of the Improvement Reward Design}

\begin{center}
\captionsetup{type=figure}
\includegraphics[width=\linewidth]{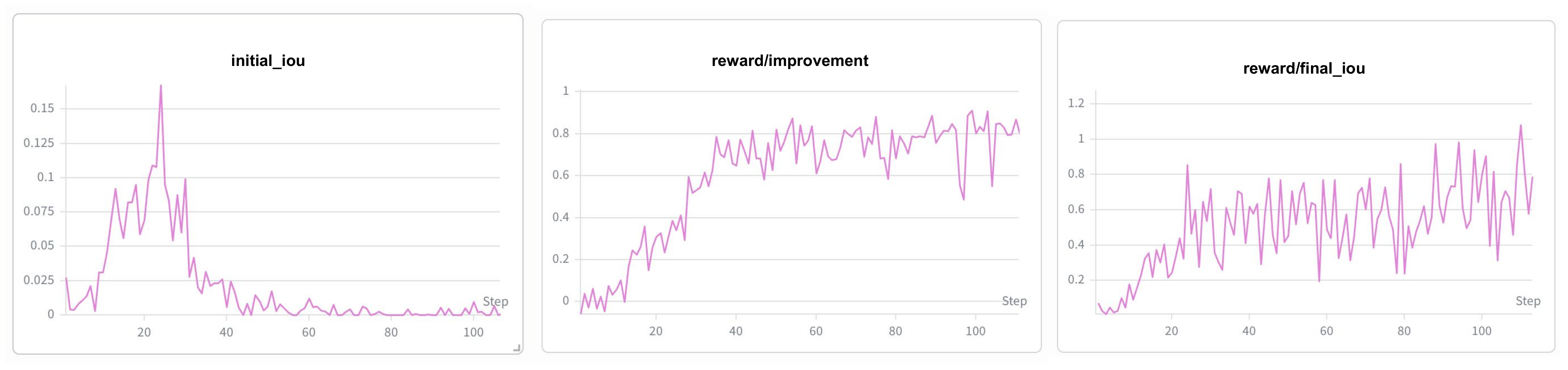}
\caption{Training curves for the ablated improvement reward variant in which the external baseline IoU is excluded, reducing the improvement term to $\max(0, R_{\text{IoU}}^{\text{final}} - R_{\text{IoU}}^{\text{initial}})$. The $x$-axis shows training steps. Panels show (left) initial-answer IoU (\texttt{initial\_iou}), (center) improvement reward (\texttt{reward/improvement}), and (right) final-answer IoU (\texttt{reward/final\_iou}). After an initial rise, the initial-answer IoU collapses toward~0 as the model learns to produce deliberately poor first predictions in order to inflate the improvement margin -- a form of reward exploitation. The full formulation (Section~\ref{sec:reward_design}), which competes against the stronger of the model's own first answer and the scaled external baseline, eliminates this collapse.}
\label{fig:reward_exploit}
\end{center}

To validate the necessity of the initial IoU reward and the external baseline reference in the improvement reward, we trained an ablated variant in which $\mathrm{IoU}_{\text{baseline}}$ is excluded, reducing the improvement reward to $\max(0, R_{\text{IoU}}^{\text{final}} - R_{\text{IoU}}^{\text{initial}})$. As shown in Figure~\ref{fig:reward_exploit}, this variant exhibits reward exploitation: after a certain number of training steps, the initial-answer IoU collapses toward zero as the model learns to produce deliberately poor first predictions in order to maximize the improvement margin. The full formulation, which competes against the stronger of the model's own first answer and the scaled external baseline, eliminates this behavior and maintains stable initial-answer quality throughout training.

As an untested alternative, $\mathrm{IoU}_{\text{baseline}}$ could instead be set to a fixed constant rather than a strong baseline model's score, which would relax the improvement bar when a high-performing external reference would otherwise make it too stringent; we did not evaluate this variant.

\section{Framework Implementation with VeRL}
\label{app:verl}

We optimize the policy with GRPO using the VeRL/EasyR1 framework. The policy is a Qwen3-VL-Instruct-4B multimodal LLM (we additionally train an 8B variant to study scaling); given an image and a query, it emits in a single generation its reasoning, the object-versus-part decision, and the grounding prediction (a bounding box and a point per target). We train the full model -- vision encoder, connector, and language model -- and freeze nothing on the policy side. Only at evaluation are the predicted box-point pairs converted to masks by a frozen SAM2-Large decoder for IoU scoring; the SAM2 receives no gradient and is not part of training. Table \ref{tab:impl_details} summarizes the full configuration.

\begin{table}[t]
\centering
\caption{Training and inference configuration.}
\label{tab:impl_details}
\small
\setlength{\tabcolsep}{6pt}
\begin{tabular}{l p{0.57\linewidth}}
\toprule
\textbf{Setting} & \textbf{Value} \\
\midrule
Policy backbone & Qwen3-VL-Instruct-4B (main); 8B for scaling \\
Mask decoder (eval only) & SAM2-Large (\texttt{sam2-hiera-large}), frozen \\
Alternate decoder & SAM3-Tracker (\texttt{facebook/sam3}, \texttt{Sam3TrackerModel}) \\
Trained parameters & Full MLLM (vision encoder + connector + LM) \\
Precision / sharding & bf16, FSDP full-shard, gradient checkpointing, CPU offload \\
RL framework & VeRL and EasyR1 (GRPO) \\
Group size (rollouts/prompt) & 4 \\
Rollout sampling & temperature 1.2, top-$p$ 1.0 \\
KL regularization & low-variance KL loss, $\beta = 1\times10^{-2}$ \\
Policy-loss clip (asymmetric) & $\epsilon_{\text{low}} = 0.2$, $\epsilon_{\text{high}} = 0.3$ \\
Reference policy & frozen snapshot of the initial policy \\
Optimizer & AdamW, lr $1\times10^{-6}$, weight decay $1\times10^{-2}$ \\
Gradient clipping & max grad norm 1.0 \\
Batch & 16 prompts/step ($\times$4 rollouts $=$ 64 responses) \\
PPO mini-batch / micro-batch & 16 / 2 (per device) \\
Training length & up to 1300 steps; checkpoint every 50 steps \\
Eval decoding & greedy, deterministic (\texttt{do\_sample=False}) \\
Compute & 4B: 4$\times$40\,GB (single node); 8B: 2$\times$141\,GB \\
Training data & VisionReasoner ($\sim$7k) + InstructPart ($\sim$1.2k) prompt--image pairs \\
\bottomrule
\end{tabular}
\end{table}

\paragraph{Optimization.} We use AdamW (learning rate $1\times10^{-6}$, weight decay $1\times10^{-2}$, gradient-norm clipping at $1.0$). Training runs in bf16 with fully-sharded data parallelism (FSDP), gradient checkpointing, and CPU offloading of parameters and optimizer state to fit the model in memory.

\paragraph{GRPO.} For each prompt we sample a group of $4$ responses (rollout temperature $1.2$, top-$p$ $1.0$) and compute group-relative advantages. The policy loss uses an asymmetric PPO clip ($\epsilon_{\text{low}}=0.2$, $\epsilon_{\text{high}}=0.3$). We regularize toward a frozen snapshot of the initial policy with a low-variance KL term applied as an auxiliary loss (coefficient $\beta = 1\times10^{-2}$). Each optimization step draws $16$ prompts ($64$ responses), with a per-device micro-batch of $2$ and gradient accumulation to the effective batch. 

\paragraph{Schedule and inference.} We train for up to $1300$ steps, save a checkpoint every $50$ steps. At evaluation we use greedy, deterministic decoding.

\paragraph{Compute and data.} The 4B model trains on $4\times40$\,GB GPUs (single node); the 8B model trains on $2\times$H200 (141\,GB) GPUs. Training uses the merged VisionReasoner\,+\,InstructPart set as described in Section~\ref{sec:datasets}. Over $1300$ steps at $16$ prompts per step, the policy processes ${\approx}20.8$k prompt samples.

\subsection{Active Visual Perception Training with VeRL}
\label{app:ap_challenges}
Recall from Section~\ref{sec:active_perception} that active visual perception grounds the self-reflective step in fresh visual evidence: after the initial localization (\texttt{<first\_answer>}), generation pauses, the predicted boxes are used to crop the corresponding image regions, and each crop is re-encoded by the vision encoder and injected back as interleaved visual tokens before the model produces its critique (\texttt{<criticism>}) and refined answer (\texttt{<answer>}). One rollout therefore proceeds as two passes with a crop-and-inject step in between, as summarized in Algorithm~\ref{alg:ap_rollout}.

\begin{algorithm}[t]
\caption{One active-perception rollout}
\label{alg:ap_rollout}
\begin{algorithmic}[1]
\Require image $I$, query $q$, policy $\pi$
\Ensure refined grounding for $q$
\State \textcolor{gray}{\# Pass 1: initial grounding}
\State $a_1 \gets \pi.\text{generate}(\text{prompt}(I, q))$ \Comment{locate $\rightarrow$ target decision $\rightarrow$ first boxes}
\State $B \gets \text{parse\_boxes}(a_1)$
\State \textcolor{gray}{\# Active perception: crop $\rightarrow$ re-encode $\rightarrow$ inject}
\State $C \gets [\,\text{crop}(I, b \ \textbf{for}\ b \in B[:\textsc{MaxCrops}]\,]$ 
\If{$C$ is empty}
    \State $C \gets [\,\text{center\_crop}(I)\,]$ \Comment{fallback so Pass~2 still gets a crop}
\EndIf
\State $F \gets \text{vision\_encoder}(C)$ \Comment{re-encode each crop at full resolution}
\State $\text{ctx} \gets \text{user\_turn}(F + \text{refine\_instructions})$ \Comment{inject as a new turn}
\State \textcolor{gray}{\# Pass 2: self-reflective step}
\State $a_2 \gets \pi.\text{generate}(\text{prompt}(I, q) + a_1 + \text{ctx})$ \Comment{critique, then refine}
\State \Return $\text{parse\_boxes}(a_2)$ \Comment{final grounding (may equal $a_1$)}
\end{algorithmic}
\end{algorithm}

\paragraph{Why this requires changes to VeRL.}
VeRL, like most RL-from-feedback frameworks, assumes the sequence it rewards and trains on was produced in a single generation. Active perception breaks that assumption: the answer arrives in two passes, with fresh images injected into the middle. We keep VeRL's outer loop intact -- sample a group of responses per query, score them, and update the policy with GRPO -- but replace the single generation call with the two-pass rollout of Algorithm \ref{alg:ap_rollout}, then stitch the two passes into one training example (first answer, injected crops, then the self-reflective step, in order) so that reward scoring and the policy update see one contiguous response. Making this work required addressing several issues, which we share below for reproducibility:

\begin{itemize}
\item \textbf{Context masking.} The injected crops sit in the middle of the combined sequence, but they are visual context the system inserted, not text the model produced. We mask these tokens out of the loss so that no gradient flows through them. The policy is trained only on tokens it generated -- its first-pass answer and its self-reflective step -- while the crops act as context.

\item \textbf{Grouping vs.\ caching.} GRPO requires the several rollouts of one query to be grouped together to compute relative advantage, but VeRL's image-feature cache assumed grouped rollouts \textit{share the same images}. This is not true for us since each rollout crops different regions. We retain the grouping while disabling the cache for these rollouts, so each rollout's own crops are encoded independently.

\item \textbf{Keeping the injected turn well-formed.} Images are only valid inside a "user" turn; injecting them into the model's own assistant (answer) turn yields malformed, off-distribution context and empty generations. We therefore close the assistant turn, open a user turn to host the crops, and reopen the assistant turn for the reflected answer.

\item \textbf{Letting the reward see the refined answer.} VeRL's reward reader assumes the answer sits at the front of the response. With a first pass and injected crops preceding it, the refined answer was being truncated. We repack the response so that all real tokens are contiguous at the front, and the reward reader then sees the full self-reflective step.

\item \textbf{Memory and length safety.} Additional images and longer sequences raise peak memory and can overflow the context window. Alongside a micro-batch of one with gradient accumulation, we cap the number of crops per sample (4), drop degenerate (sliver-thin) crops, and, if a self-reflective prompt would still overflow, rebuild it without crops and cleanly drop that example from the update.
\end{itemize}

Together, these changes turn VeRL's one-shot loop into a genuine perceive, act, re-perceive, and refine cycle, while leaving the standard single-pass training path unchanged.

\section{Evaluation Protocol}
\label{app:eval}

Our task is visual grounding: given an image and a query naming an object or an object part, the model localizes the referenced region and produces a segmentation mask. For object parts, queries take the form ``\textit{object's part}'' (e.g., ``dog's ear''), reflecting the object--part relationship our method targets. Following the standard grounding setting, we query only objects and parts known to be present in the image, obtain the predicted mask, and measure its overlap with the ground-truth mask. Throughout the paper we report generalized IoU (gIoU), the mean of the per-query IoU over $N$ evaluation queries with predicted masks $P_i$ and ground-truth masks $G_i$,
\begin{equation}
\text{gIoU} = \frac{1}{N}\sum_{i=1}^{N} \frac{|P_i \cap G_i|}{|P_i \cup G_i|}.
\end{equation}
In Section~\ref{app:extended_results} we additionally report cumulative IoU (cIoU), which aggregates intersections and unions across the entire dataset before dividing,
\begin{equation}
\text{cIoU} = \frac{\sum_{i=1}^{N} |P_i \cap G_i|}{\sum_{i=1}^{N} |P_i \cup G_i|}.
\end{equation}
gIoU weights every sample equally regardless of region size, whereas cIoU is biased toward larger regions, since bigger masks contribute proportionally more to the aggregate. Because part regions are typically small and vary widely in scale, gIoU is our primary metric. We report cIoU for completeness. 

\section{Extended Results (gIoU and cIoU)}
\label{app:extended_results}
Table \ref{tab:extended_results} shows gIoU and cIoU for our headline method along with a subset of the baselines.

\begin{table}[t]
\centering
\caption{Extended results with both gIoU and cIoU (reported as gIoU\,/\,cIoU) on part-grounding benchmarks. $^\dagger$InstructPart is evaluated on its test split; our RL training uses the InstructPart train split.}
\label{tab:extended_results}
\small
\setlength{\tabcolsep}{3pt}
\begin{tabular}{lcccc}
\toprule
& \multicolumn{3}{c}{\textbf{Parts}} & \textbf{Objects} \\
\cmidrule(lr){2-4} \cmidrule(lr){5-5}
\textbf{Method} & \textbf{InstructPart}$^\dagger$ & \textbf{PascalPart} & \textbf{PartImgNet} & \textbf{Pascal-Obj} \\
\midrule
LISA-7B & 43.26\,/\,48.39 & 13.82\,/\,26.29 & 29.91\,/\,40.74 & 83.50\,/\,88.40 \\
Sa2VA-4B & 50.15\,/\,49.95 & 14.67\,/\,19.49 & 38.13\,/\,45.78 & 77.02\,/\,78.77 \\
PixelLM-7B & 44.41\,/\,50.97 & 16.57\,/\,24.51 & 35.25\,/\,42.20 & 81.71\,/\,86.89 \\
UniPixel-3B & 59.68\,/\,61.07 & 30.64\,/\,45.01 & 46.18\,/\,51.84 & 85.54\,/\,89.31 \\
ClipSeg (OV-PARTS) & 31.85\,/\,34.09 & 12.51\,/\,29.10 & 33.45\,/\,42.93 & 70.51\,/\,78.91 \\
\midrule
Qwen3-VL-4B+SAM2 & & & & \\
(OP-HRG Prompt) & 31.45\,/\,43.04 & 12.83\,/\,21.32 & 21.95\,/\,35.44 & 36.91\,/\,48.67 \\
\rowcolor{blue!8} Ours & \textbf{75.56}\,/\,\textbf{77.01} & \textbf{38.59}\,/\,\textbf{51.24} & \textbf{56.87}\,/\,\textbf{60.26} & \textbf{87.50}\,/\,\textbf{89.51} \\
\bottomrule
\end{tabular}
\end{table}

\section{Qualitative Examples}
\label{app:qualitative}

\subsection{Examples on different benchmarks}
Figures~\ref{fig:qual_instructpart_1}--\ref{fig:qual_partimagenet_2} show representative predictions across our three part benchmarks: InstructPart (Figs.~\ref{fig:qual_instructpart_1}--\ref{fig:qual_instructpart_3}), PascalPart (Figs.~\ref{fig:qual_pascalpart_1}--\ref{fig:qual_pascalpart_2}), and PartImageNet (Figs.~\ref{fig:qual_partimagenet_1}--\ref{fig:qual_partimagenet_2}). For each example we show the query, the ground-truth mask, the predicted object hint, the initial and final answers with their masks, and the raw model output including the reasoning trace and self-critique. These illustrate how the model decomposes a part query, anchors on the parent object, and produces a tight part localization. All examples use a trained Qwen3-VL-Instruct-4B with a frozen SAM2 decoder.

\section{Additional Ablations}
\label{app:more_ablations}

\subsection{Backbone Selection: Instruct vs.\ Think Variant}
\label{sec:instruct_vs_think}

We compare the two Qwen3-VL-4B variants -- Instruct and Think -- under our OP-HRG structured prompt without RL training to motivate our backbone choice. As shown in Table~\ref{tab:instruct_vs_think}, the Instruct variant consistently outperforms the Think variant by a wide margin across all three benchmarks. This aligns with the Qwen3-VL technical report \cite{bai2025qwen3}, which reports stronger grounding performance for the Instruct variants on RefCOCO. The Think variant's extended internal reasoning appears to interfere with reliable adherence to the structured OP-HRG output format, whereas the instruction-tuned variant is better suited to following the fixed sequence of tagged reasoning and localization steps of OP-HRG. We therefore adopt Qwen3-VL-Instruct-4B as our primary backbone throughout all experiments.

\begin{table}[ht]
\centering
\caption{Vanilla (no RL) part-grounding performance (gIoU) of Qwen3-VL-4B variants under the OP-HRG structured prompt. Both use SAM2 as the mask decoder.}
\label{tab:instruct_vs_think}
\small
\begin{tabular}{lccc}
\toprule
\textbf{Dataset} & \textbf{Instruct-4B} & \textbf{Think-4B} & $\boldsymbol{\Delta}$ \textbf{(Think--Inst)} \\
\midrule
InstructPart$^\dagger$ & 31.45 & 7.43 & $-$24.02 \\
PascalPart & 12.83 & 6.30 & $-$6.53 \\
PartImageNet & 21.95 & 10.61 & $-$11.34 \\
\bottomrule
\end{tabular}
\end{table}

\subsection{The Single-Step-Prompt Baseline and the Reasoning-Execution Gap}
\label{sec:bare_prompt}

A natural concern is whether our gains stem from the reasoning-guided protocol or merely from the base model's raw localization capability. We therefore evaluate the same Qwen3-VL-Instruct-4B under a plain ``single-step'' prompt that requests only bounding boxes and representative points -- no hierarchy, no first answer, no self-critique, no refinement -- and compare it with the base model under our full structured OP-HRG protocol and with our RL-trained model. Results are shown in Table~\ref{tab:bare_prompt}.

\begin{table}[ht]
\centering
\caption{Effect of prompt structure and RL training on Qwen3-VL-Instruct-4B + SAM2 (gIoU). The plain prompt requests only boxes and points with no structured reasoning.}
\label{tab:bare_prompt}
\small
\begin{tabular}{lcc}
\toprule
\textbf{Condition} & \textbf{InstructPart}$^\dagger$ & \textbf{PartImageNet} \\
\midrule
Plain prompt & 72.39 & 50.20 \\
OP-HRG prompt (no RL) & 31.45 & 21.95 \\
OP-HRG prompt + RL (Ours) & \textbf{75.56} & \textbf{56.87} \\
\bottomrule
\end{tabular}
\end{table}

The untrained model fails reasoning-guided grounding at two levels. First, it cannot reliably produce the structured protocol: prompted zero-shot with our six-step format, it emits valid, parseable output only 52.7\% of the time on InstructPart and 56.6\% on PartImageNet; the remaining responses are malformed and score zero. Second, even on the subset that does parse correctly, its localization is weaker than under the plain prompt (gIoU 59.7 vs.\ 72.4 on InstructPart; 38.8 vs.\ 50.2 on PartImageNet). The structured scaffolding meant to guide grounding instead degrades it when the model has not learned to execute it.

Training closes both layers of this gap. After RL, the model emits valid structured output on 100\% of queries and its localization surpasses the minimal prompt ceiling: +3.17 gIoU in-domain (InstructPart, 72.39 $\to$ 75.56) and +6.67 on the harder cross-dataset zero-shot split (PartImageNet, 50.20 $\to$ 56.87). Since the simple-prompt baseline is itself well-formed (99.7\% parseable) and a capable localizer, this margin cannot be attributed to formatting artifacts -- it isolates the value that structured reasoning adds once the model has learned to execute it. The gain is large on out-of-distribution parts, demonstrating the effectiveness of  hierarchical reasoning and self-reflection.

\section{Baseline Provenance}
\label{app:provenance}

To ensure a transparent comparison, we document the source of every baseline number reported in the paper. For our main part-grounding results (Table~\ref{tab:zero_shot_part}), \emph{all} values -- for every baseline and for our own models -- were computed by us under a single, consistent evaluation protocol, rather than copied from different sources with differing settings. We will release the full evaluation code and configurations to allow reproduction of every value in Table \ref{tab:zero_shot_part}.

For the referring-comprehension and reasoning-segmentation tables, several numbers are drawn from prior work, since those benchmarks have well-established results. We report each baseline from its primary source wherever the corresponding setting is available, and otherwise attribute the value to the work from which it was obtained.

\subsection{Referring Expression Comprehension (Table~\ref{tab:refcoco})}
\begin{itemize}
  \item \textbf{MDETR}~\cite{kamath2021mdetr}: from the original paper
    (ResNet-101, trained on RefCOCO/+/g).
  \item \textbf{GLIP-T}~\cite{li2022grounded}: from Grounding
    DINO~\cite{liu2024grounding}, zero-shot row (no RefCOCO training); marked
    \textbf{ZS}.
  \item \textbf{G-DINO-T}~\cite{liu2024grounding}: from Grounding DINO, RefCOCO
    fine-tuned row.
  \item \textbf{Shikra-7B}~\cite{chen2023shikra},
    \textbf{InternVL2-8B}~\cite{chen2024internvl},
    \textbf{VisionReasoner-7B}~\cite{visionreasoner}: from their
    respective papers.
  \item \textbf{Qwen3-VL-4B}~\cite{bai2025qwen3}: the Qwen3-VL technical report \cite{bai2025qwen3} does
    not explicitly report the test-split value, so we list the values
    re-evaluated by EGM (Instruct variant) \cite{zhan2026scaling}.
  \item \textbf{Qwen2.5-VL-7B}~\cite{bai2025qwen25vl}: from the original
    technical report.
\end{itemize}

\subsection{Reasoning Segmentation (Table \ref{tab:reasonseg})}
\begin{itemize}
  \item \textbf{ReLA}~\cite{liu2023gres}: ReasonSeg evaluation
    from~\cite{yang2023lisapp}.
  \item \textbf{LISA-7B}~\cite{lisa},
    \textbf{VisionReasoner-7B}~\cite{visionreasoner}, \textbf{GenSeg-R1-4B} \cite{hegde2026genseg}: from their
    respective papers.
  \item \textbf{Seg-Zero-7B} and \textbf{Qwen2.5-VL-7B}: from
    VisionReasoner~\cite{visionreasoner}, which evaluates all MLLM
    baselines under a unified box-to-mask protocol (boxes prompted into
    SAM2~\cite{sam2}), matching the protocol used for our model.
\end{itemize}

\clearpage
\begin{center}
\vspace*{\fill}
\includegraphics[width=\linewidth]{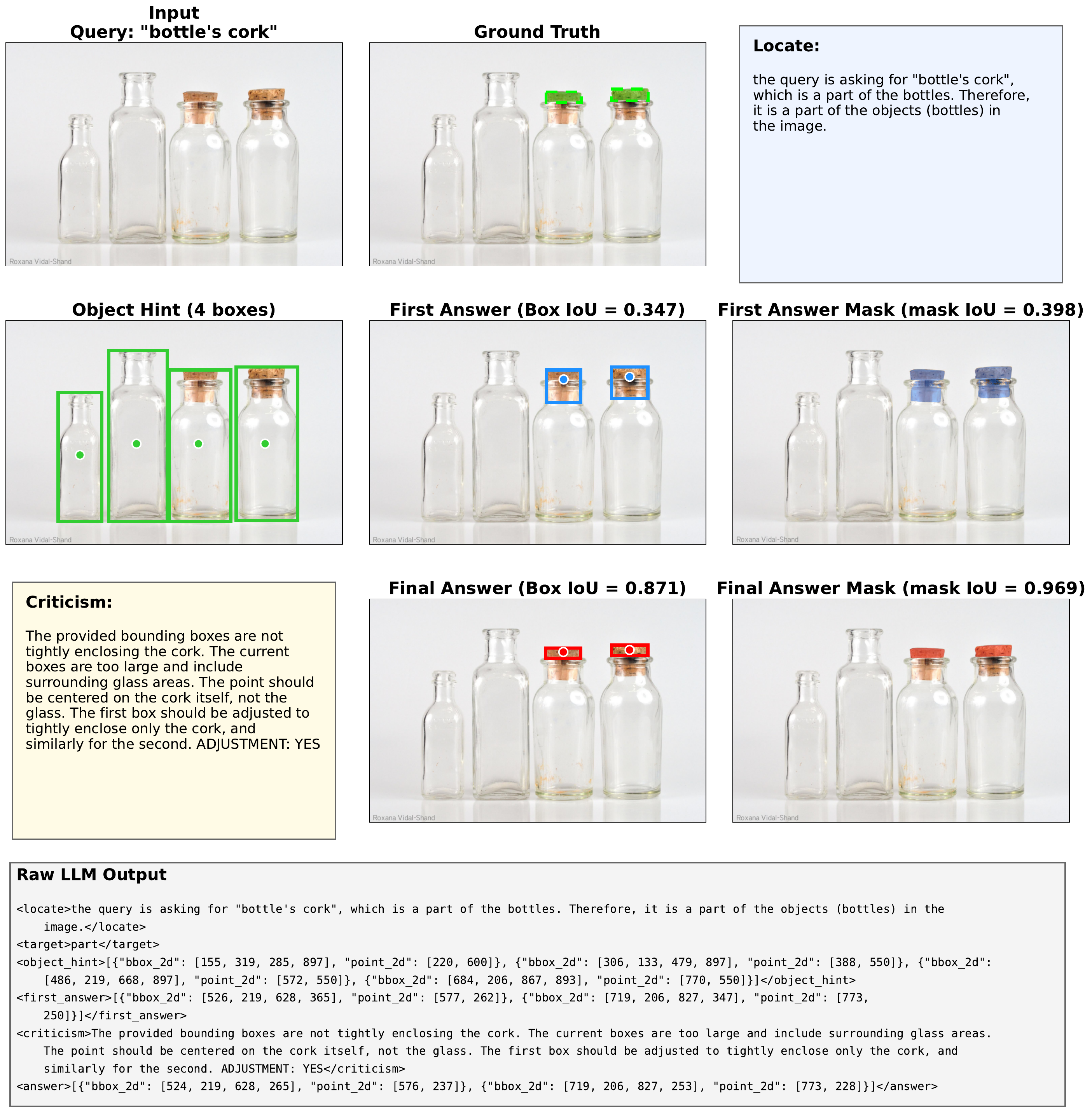}
\captionsetup{hypcap=false}\captionof{figure}{Qualitative example on InstructPart.}
\label{fig:qual_instructpart_1}
\vspace*{\fill}
\end{center}

\clearpage
\begin{center}
\vspace*{\fill}
\includegraphics[width=\linewidth]{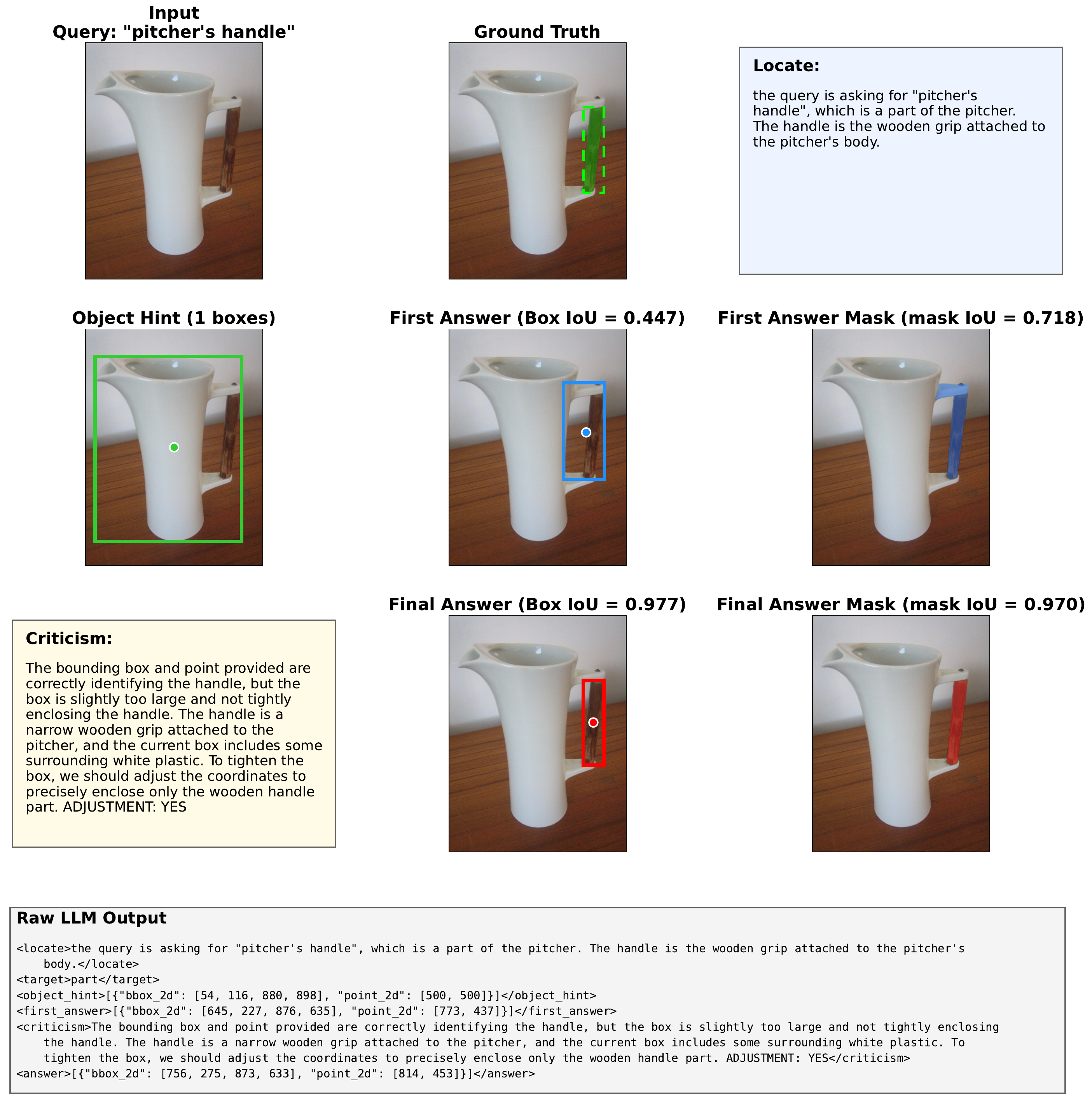}
\captionsetup{hypcap=false}\captionof{figure}{Qualitative example on InstructPart.}
\label{fig:qual_instructpart_2}
\vspace*{\fill}
\end{center}

\clearpage
\begin{center}
\vspace*{\fill}
\includegraphics[width=\linewidth]{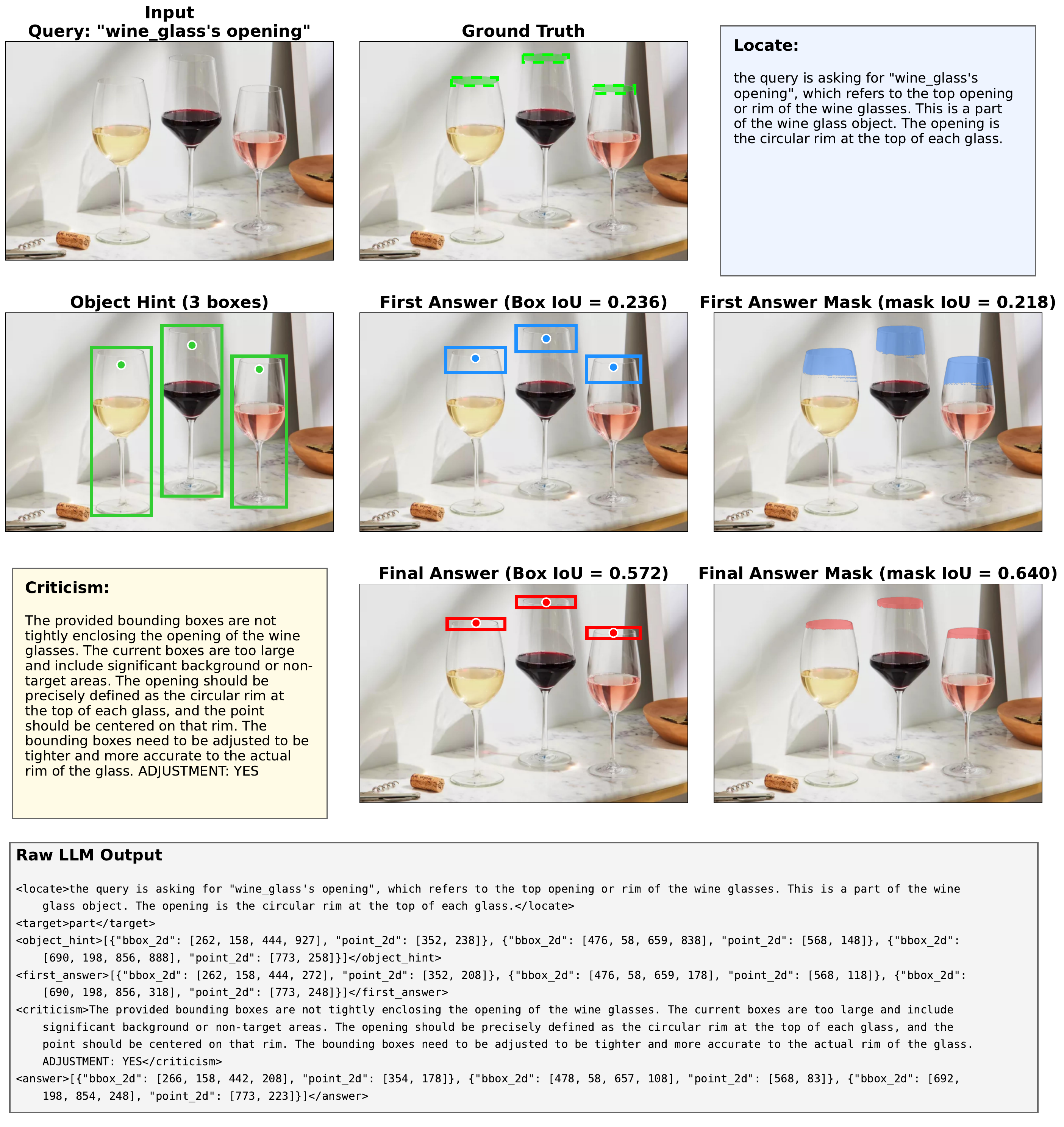}
\captionsetup{hypcap=false}\captionof{figure}{Qualitative example on InstructPart.}
\label{fig:qual_instructpart_3}
\vspace*{\fill}
\end{center}

\clearpage
\begin{center}
\vspace*{\fill}
\includegraphics[width=\linewidth]{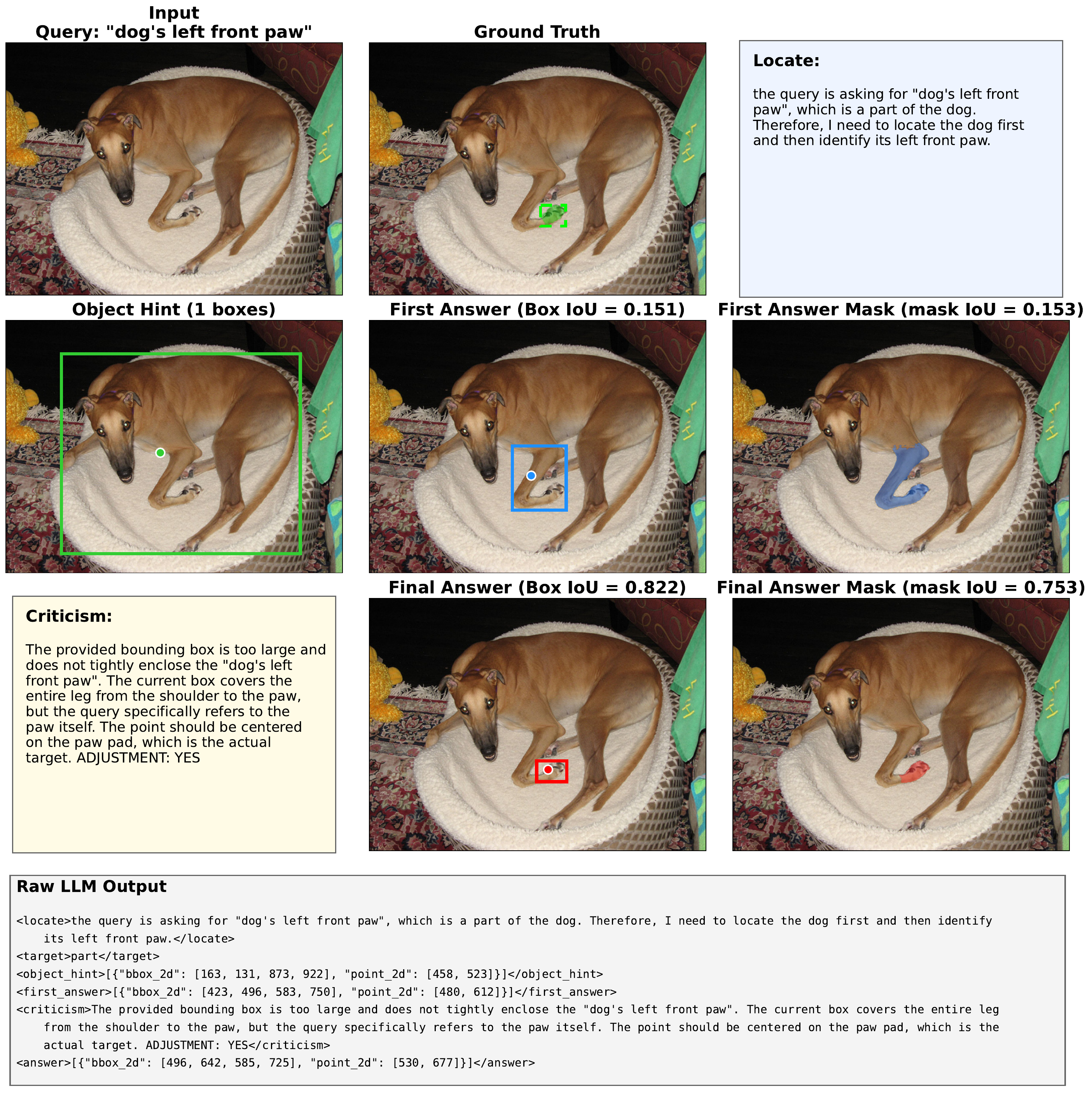}
\captionsetup{hypcap=false}\captionof{figure}{Qualitative example on PascalPart.}
\label{fig:qual_pascalpart_1}
\vspace*{\fill}
\end{center}

\clearpage
\begin{center}
\vspace*{\fill}
\includegraphics[width=\linewidth]{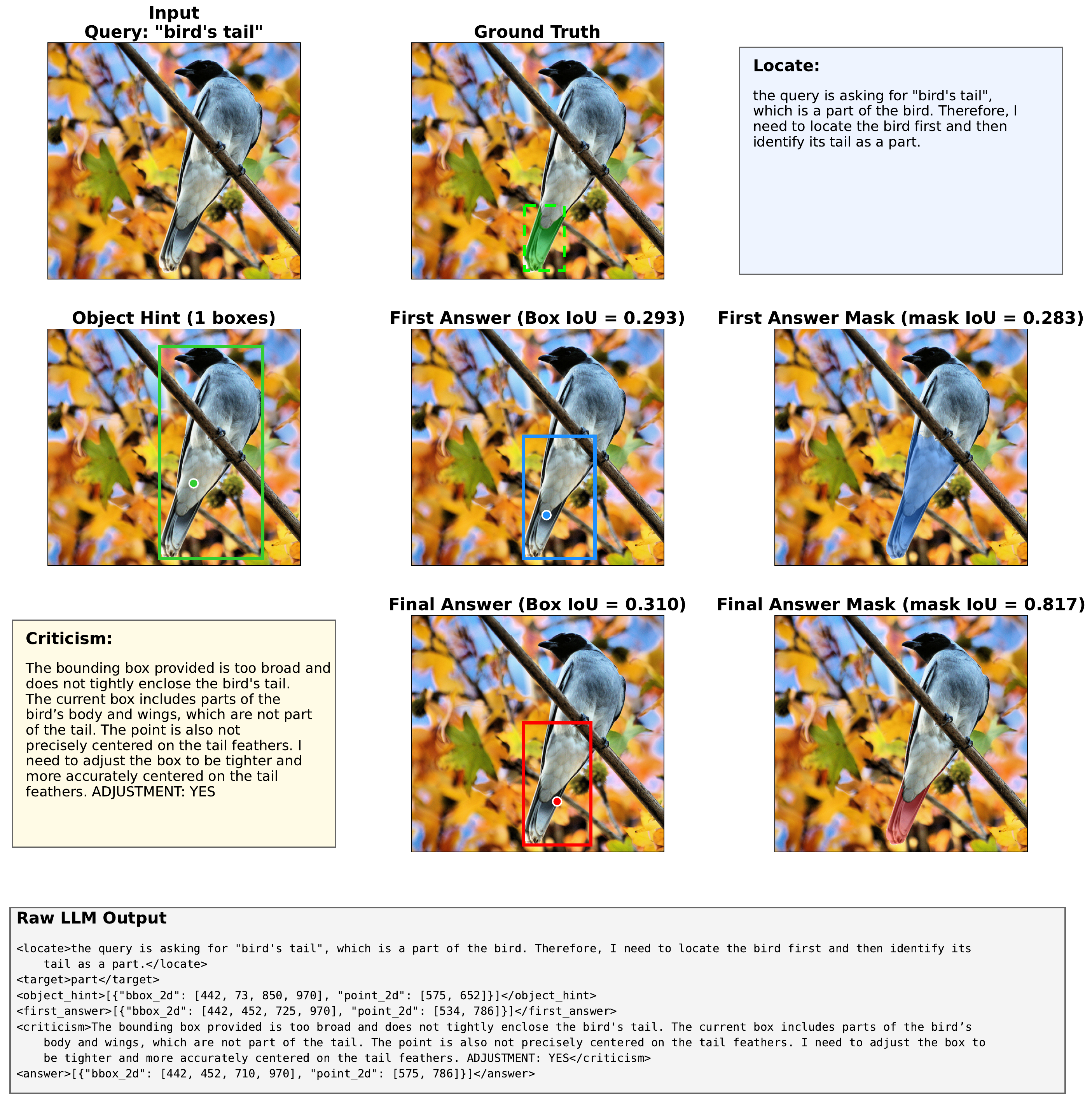}
\captionsetup{hypcap=false}\captionof{figure}{Qualitative example on PascalPart.}
\label{fig:qual_pascalpart_2}
\vspace*{\fill}
\end{center}

\clearpage
\begin{center}
\vspace*{\fill}
\includegraphics[width=\linewidth]{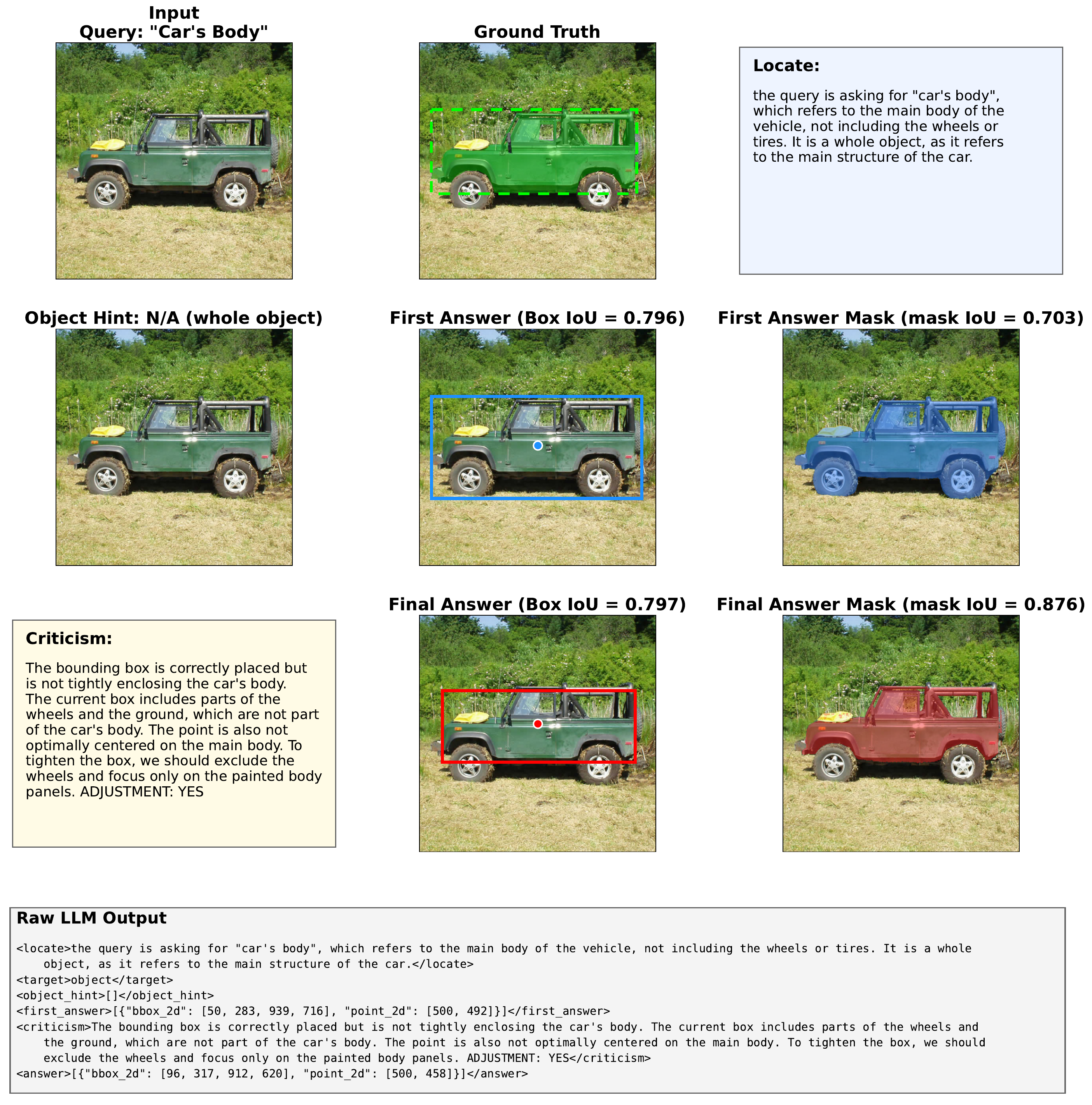}
\captionsetup{hypcap=false}\captionof{figure}{Qualitative example on PartImageNet.}
\label{fig:qual_partimagenet_1}
\vspace*{\fill}
\end{center}

\clearpage
\begin{center}
\vspace*{\fill}
\includegraphics[width=\linewidth]{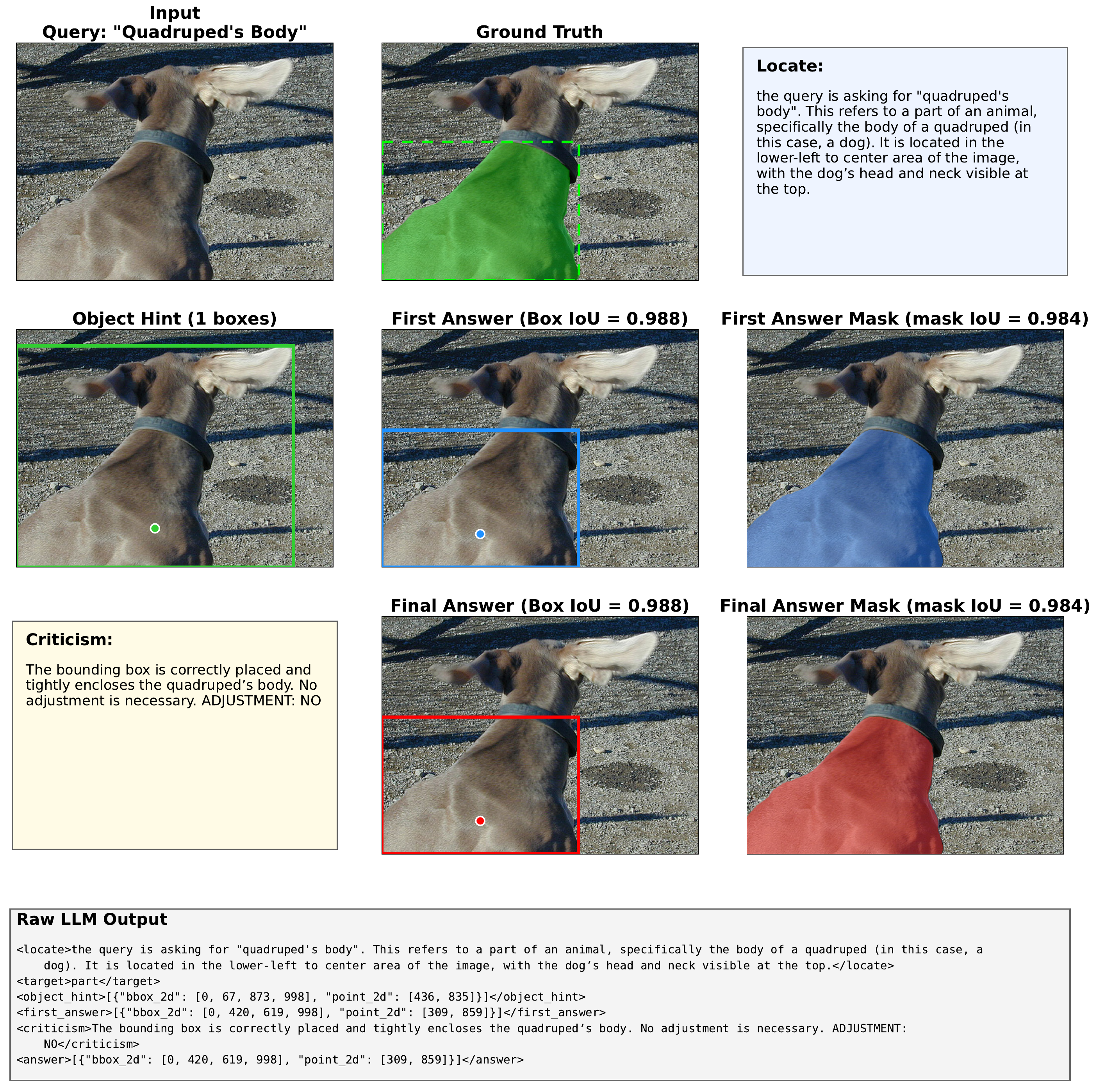}
\captionsetup{hypcap=false}\captionof{figure}{Qualitative example on PartImageNet.}
\label{fig:qual_partimagenet_2}
\vspace*{\fill}
\end{center}

\clearpage

\end{document}